\documentclass[10pt,twocolumn,letterpaper]{article}
 
\usepackage{cvpr}              

\usepackage{multirow}
\usepackage{multicol}
\usepackage{algorithm}
\usepackage{bm}
\usepackage{algpseudocode}
\usepackage[dvipsnames]{xcolor}
\usepackage{float}
\usepackage{array} 
\definecolor{mygreen}{RGB}{112, 173, 71}
\definecolor{myorange}{RGB}{237, 125, 49}
\definecolor{xigreen}{RGB}{0, 100, 0}
\newcommand{\algorithmicoutput}{\textbf{Return:}}
\newcommand{\OUTPUT}{\item[\algorithmicoutput]}

\newcommand{\methodname}{2DMamba}
\newcommand{\methodnameMIL}{2DMambaMIL}

\newcommand{\beginsupplement}{%
        \setcounter{table}{0}
        \renewcommand{\thetable}{S\arabic{table}}%
        \setcounter{figure}{0}
        \renewcommand{\thefigure}{S\arabic{figure}}%
     }

\renewcommand{\thealgorithm}{}

\definecolor{cvprblue}{rgb}{0.21,0.49,0.74}
\usepackage[pagebackref,breaklinks,colorlinks,allcolors=cvprblue]{hyperref}

\def\paperID{5815} 
\def\confName{CVPR}
\def\confYear{2025}

\title{2DMamba: Efficient State Space Model for Image Representation with Applications on Giga-Pixel Whole Slide Image Classification}

\newcommand*\samethanks[1][\value{footnote}]{\footnotemark[#1]}
\newcommand{\cbm}[1]{{\color{BurntOrange} \bm{#1}}}

\author{
Jingwei Zhang$^{1}$\thanks{These authors contributed equally to this paper.},  Anh Tien Nguyen$^{2}$\samethanks[1], Xi Han$^{1}$\samethanks[1], Vincent Quoc-Huy Trinh$^{5}$, Hong Qin$^{1}$, \\
Dimitris Samaras$^{1}$, Mahdi S. Hosseini$^{3,4}$\\
$^1$Stony Brook University, Stony Brook, NY, USA 
$^2${Korea University, Seoul, South Korea} \\
$^3$Concordia University, Montreal, Canada 
$^4$Mila–Quebec AI Institute, Montreal, Canada \\
$^5$University of Montreal Hospital Center, Montreal, Canada 
}

\begin{document}
\maketitle


\begin{abstract}
Efficiently modeling large 2D contexts is essential for various fields including Giga-Pixel Whole Slide Imaging (WSI) and remote sensing. Transformer-based models offer high parallelism but face challenges due to their quadratic complexity for handling long sequences. Recently, Mamba introduced a selective State Space Model (SSM) with linear complexity and high parallelism, enabling effective and efficient modeling of wide context in 1D sequences. However, extending Mamba to vision tasks, which inherently involve 2D structures, results in spatial discrepancies due to the limitations of 1D sequence processing. On the other hand, current 2D SSMs inherently model 2D structures but they suffer from prohibitively slow computation due to the lack of efficient parallel algorithms. In this work, we propose \methodname, a novel 2D selective SSM framework that incorporates the 2D spatial structure of images into Mamba, with a highly optimized hardware-aware operator, adopting both spatial continuity and computational efficiency. We validate the versatility of our approach on both WSIs and natural images. Extensive experiments on 10 public datasets for WSI classification and survival analysis show that \methodname~improves up to $2.48\%$ in AUC, $3.11\%$ in F1 score, $2.47\%$ in accuracy and $5.52\%$ in C-index. Additionally, integrating our method with VMamba for natural imaging yields $0.5$ to $0.7$ improvements in mIoU on the ADE20k semantic segmentation dataset, and $0.2\%$ accuracy improvement on ImageNet-1K classification dataset. Our code is available at \href{https://github.com/AtlasAnalyticsLab/2DMamba}{https://github.com/AtlasAnalyticsLab/2DMamba}.
\end{abstract}

\section{Introduction}
\label{sec:intro}
    \begin{figure}[t] 
      \centering
      \includegraphics[width=1\linewidth]{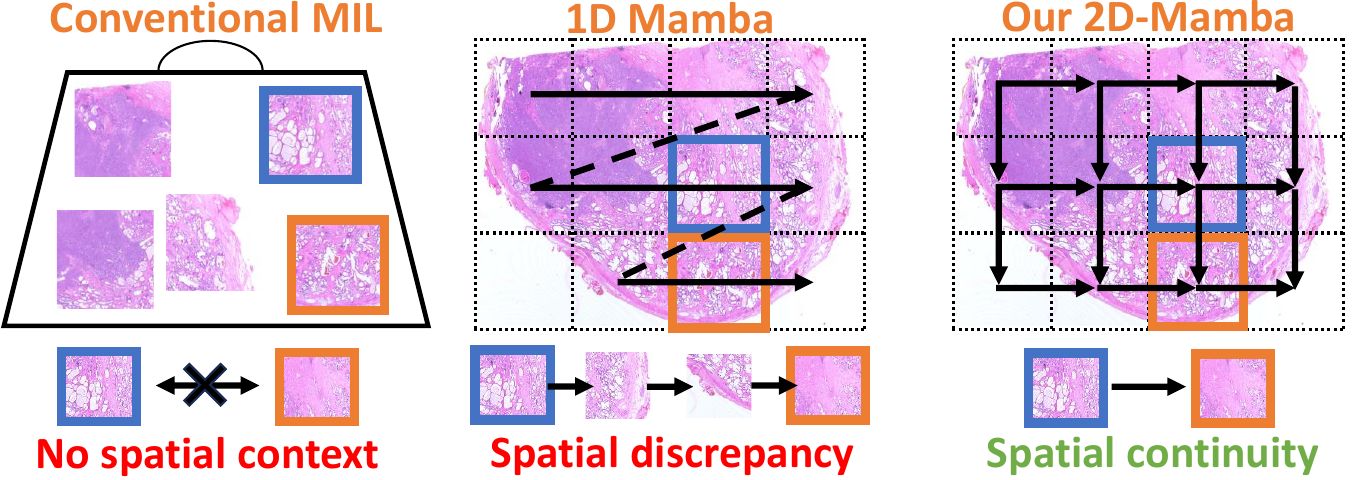}
      \caption{
        \textbf{Left: Conventional MIL Bagging} of patches adopts \textit{no spatial context}. \textbf{Middle: 1D Mamba-based methods} flatten a WSI into a 1D sequence and lose the 2D structure. The adjacent \textcolor{blue}{blue} and \textcolor{orange}{orange} patches are far away in the sequence. We call this ``\textit{spatial discrepancy}''. \textbf{Right: \methodname} processes a WSI in a 2D manner, preserving 2D structure. The \textcolor{blue}{blue} and \textcolor{orange}{orange} patches maintains adjacent in the sequence. We call it ``\textit{spatial continuity}''.}
      \label{fig:comparison}
    \end{figure}
    Efficient understanding of large contexts over a 2D visual domain is crucial across fields such as medical imaging and remote sensing \cite{han2023survey,Chen_2023,Chen_2024}. While recurrent neural networks (RNNs) \cite{rnn} can model wide context in long sequences, their sequential nature limits parallelism, making them unable to fully utilize GPUs. As a remedy, Transformers \cite{transformer, vit}, possessing a high capacity of parallelism, became the mainstream to model long sequences, albeit with quadratic complexity. As a solution, Mamba \cite{mamba, mamba2}, benefiting from both linear-time complexity and parallelism, has emerged as a promising avenue. Mamba is a State Space Model (SSM), a mathematical framework used in control theory to capture dynamic interactions between state variables \cite{ssm_app_1,ssm_app_2,ssm_app_3}. It introduces a selective mechanism that enhances the flexibility of SSMs, allowing them to capture essential information and ignore irrelevant context. \\
    \indent Mamba was proposed for language modeling by only processing 1D sequences and extended to vision domain \cite{visionmamba,vmamba}. Due to the 2D nature of vision tasks, Mamba-based approaches for natural images have attempted to incorporate 2D image structures by adopting various formulations for flattening 2D images into 1D sequences or by scanning images in multiple directions simultaneously \cite{visionmamba,vmamba,plainmamba,pei2024efficientvmamba}. However, these methods \textbf{all} flatten 2D images into 1D sequences, which inevitably leads to the loss of spatial structure in at least one direction. This limitation persists regardless of the specific flattening strategy employed, resulting in suboptimal performance. We call this issue as ``\textit{spatial discrepancy}'' illustrated in Fig. \ref{fig:comparison}. \\
    \indent One alternative solution is 2D SSMs to maintain spatial continuity of 2D structures \cite{2dssm, s4mil}. However, unlike the highly parallel Mamba architecture, achieving parallel implementation for these methods still remains a big challenge. As a consequence, similar to traditional RNNs, these methods experience very slow computation, making them almost impractical. Moreover, they lack Mamba's selective mechanism, resulting in suboptimal performance.\\
    \indent Beyond applications in general vision tasks, Mamba finds a great potential in computational pathology, particularly for the classification of Giga-pixel Whole Slide Images (WSIs), known as the gold standard for cancer diagnosis \cite{hosseini2024survey, zhu2017wsisa, sertel2009computer,gpc,tqx,vleer}. WSIs are high-resolution images of tissue samples, often reaching up to $100,000 \times 100,000$ pixels at 40x magnification, making them extremely large and rich in spatial detail. Due to their enormous size, WSIs are typically analyzed in a Multiple Instance Learning (MIL) manner: conventional bag-based MIL methods convert a WSI into a ``bag'' of instances (patches) that are usually aggregated independently, neglecting the spatial awareness among patches \cite{abmil,clam,zhang2023prompt}. In contrast, Mamba-based methods treat the WSI as a sequence of patches \cite{s4mil,mambamil}, enabling more effective information aggregation and potentially enhancing diagnostic insights. However, they still flatten 2D images into 1D sequences, and \textit{spatial discrepancies} persist, as illustrated in Fig.~\ref{fig:comparison}. Given that cells interact with each other in a coordinated manner across all directions, scanning in multiple directions \cite{vmamba,groupmamba} does not accurately model the complexity of cell-to-cell interactions.\newline  
    \indent We propose a novel framework \textbf{\methodname} to overcome the limitations posed by the 1D nature of Mamba and the sequential nature of 2D SSMs. In summary, 
    \begin{itemize}
        \item We propose a \textit{2D selective State Space Model architecture} which directly scans a 2D image without first flattening it into a 1D sequence. It maintains the 2D structure of images and we call this ``\textit{spatial continuity}'' (Fig. \ref{fig:comparison}).
        \item We propose a novel \textit{hardware-aware 2D selective scan operator} to extend the 1D Mamba parallelism into 2D.
        \item We validate the versatility of our architecture by implementing it on two very different domains, Giga-pixel WSIs for MIL aggregation and $224 \times 224$ natural images.
    \end{itemize}
    To our best knowledge, \methodname~is the \textbf{first} intrinsic 2D Mamba method with an efficient parallel algorithm.
    Extensive experiments on $10$ public datasets for WSI classification and survival analysis show that our method achieves a relative improvement of up to $2.48\%$ better AUC, $3.11\%$ better F1, $2.47\%$ better Accuracy, and $5.52\%$ better C-index.
    We also integrate our scanning approach into the SOTA method, VMamba~\cite{vmamba}. We outperform the SOTA method by 0.5 to 0.7 in mIoU on the ADE20k semantic segmentation dataset and max the SOTA method by $0.2\%$ in accuracy on the ImageNet-1K classification dataset.

\section{Related work}
\label{sec:related_work}    
    \textbf{State Space Model (SSM)}. SSM \cite{ssm} is an effective sequence model that represents systems evolving over time by defining hidden states and their transitions, which makes it particularly useful for capturing dynamic temporal behavior in sequential data. Gu et al. \cite{gu2021combining} unified RNNs, temporal convolutions, and neural differential equations with a linear state-space layer and demonstrated the potential of SSM-based models with the HiPPO initialization. S4 \cite{gu2021efficiently} proposed to normalize the parameter matrices into a diagonal structure. 2D-SSM \cite{2dssm} adopted Roesser's 2D-SSM recursion \cite{2dssm_R} and applied it to 2D images. However, prior to Mamba, all these SSM methods suffered from slow training speed as the sequential dependency of states makes an efficient parallel algorithm very difficult.\\
    \indent\textbf{Mamba}. To accelerate SSM methods, Mamba \cite{mamba} incorporated a selective mechanism that makes the model parameters input-dependent and eliminates long dependencies by forgetting less relevant states. It also introduced a hardware-aware algorithm that drastically accelerates state computation. It was originally applied to language tasks and Vim \cite{visionmamba} introduced a Vision Mamba block that uses two independent selective SSMs for bidirectional aggregation of information in the vision domain. PlainMamba \cite{plainmamba} used a 4-directional selective scan and adopted a more spatially continuous scan path. Similarly, VMamba \cite{vmamba} and GroupMamba \cite{groupmamba} also utilized this 4-directional scan in a hierarchical network and optimized the network structure. However, the current formulations of these Mamba-based models are still limited to 1D.\\
    \indent
    \textbf{Application of MIL in WSI classification}. MIL methods are the mainstream on WSI classifications. It aggregates embedded features from a WSI for slide-level representation. AB-MIL \cite{abmil} introduced an attention-based aggregation, where the attention values were learned by a neural network. Based on that, CLAM \cite{clam} proposed a multi-branch pooling mechanism to improve the performance. DSMIL \cite{dsmil} employed multi-scale patch features in a dual-stream architecture. TransMIL \cite{shao2021transmil} introduced multi-head self-attention layers to capture both morphological and spatial relationships and used nyström Attention \cite{xiong2021nystromformer} to alleviate the quadratic complexity of self-attention. DTFD-MIL \cite{zhang2022dtfd} introduced a double-tier MIL framework by incorporating pseudo-bags. Recently, S4-MIL \cite{s4mil} and MambaMIL \cite{mambamil} used Mamba for better capturing the information in long patch sequences. However, these works still fail to fully utilize the 2D spatial information of a WSI.
    
\begin{figure*}[ht] 
      \centering
      \includegraphics[width=1\linewidth]{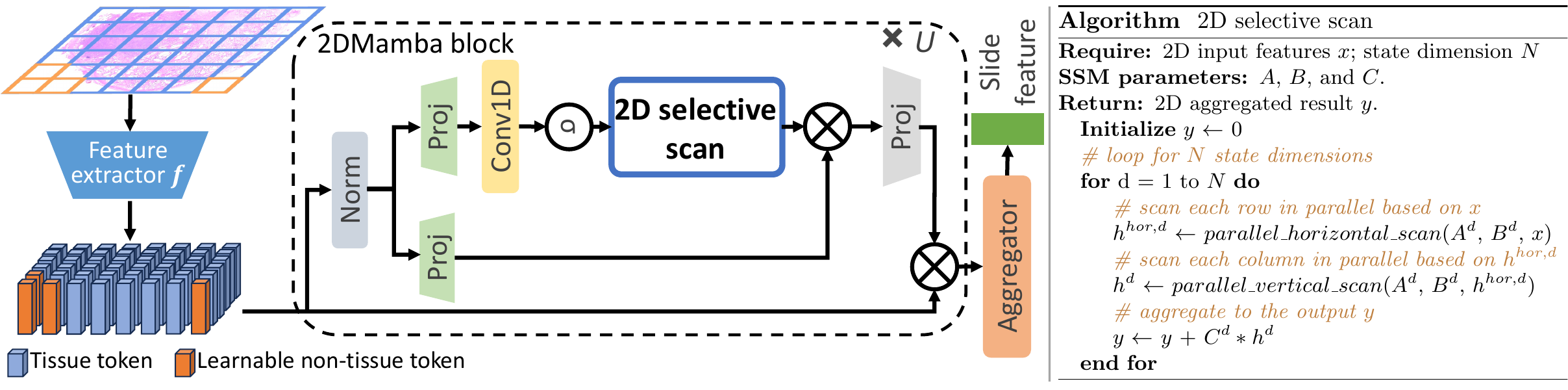}
      \caption{\textbf{Left}: The overall architecture of \methodnameMIL~for WSI representation. An input WSI is first tiled into patches and these patches are embedded by a feature extractor into a 2D features map. Non-tissue regions are padded with the learnable token to maintain the 2D spatial relationships. The 2D feature map is then fed to $U$ layers of 2D-Mamba blocks, where the key difference, compared with vanilla Mamba block, is our \textbf{2D selective scan} module. \textbf{Right}: Our 2D selective scan algorithm. It performs parallel horizontal scan and parallel vertical scan for each state dimension $d$ independently. Parameter $C$ then aggregates $N$ state dimensions into a single dimension output $y$.
      }
      \label{fig:2dmamba}
\end{figure*}
    
\section{Method}
\label{sec:method}
We present our \methodname~designed for effectiveness and efficiency, and an associated framework for WSI representation: \methodnameMIL. 
\subsection{SSM in Mamba and 1D selective scan} 
We revisit SSM, a mathematical model used to capture the behavior of dynamic systems. SSMs are designed as a \textit{function-to-function} for continuous systems and after discretization it becomes a \textit{sequence-to-sequence} model:
    \begin{align}
        \label{eq_discrete_ssm_state}
        h_t^d &= \bar{A}^dh_{t-1} + \bar{B}^dx_t^d  \\
        y_t &= Ch_t = \sum_{d=1}^{N} C^d h_t^d \ ,
        \label{equ:mamba_y}
    \end{align}       
    where $h_t^d$ is the latent state at time $t$, $y_t$ is the output, and $d\in\{1,2,\dots, N\}$ is the state dimension. The parameters $\bar{A}^d$ and $\bar{B}^d$ are time-invariant, making them non-adaptive to the input. This design limits the context-aware ability of SSMs to handle long sequence inputs.
    
    The vanilla Mamba block \cite{mamba} introduces a selective mechanism to allow the SSM to dynamically adapt to the input context. This aggregates important input into the hidden state while unimportant input can be ignored. 
    Mathematically, the parameters are formulated as functions of the input $x_t$:
    \begin{equation}
    \begin{aligned}
    \bar{A}_t^d &= \exp(\Delta_t A^d) \ ,
    \quad
    &\bar{B}_t^d &= \Delta_t B^d(x_t) \ , \\
    C_t^d &= C^d(x_t) \ ,
    \quad
    &\Delta_t &= \mathrm{softplus}(\Delta(x_t)) \ ,
    \label{equ:ssm} 
    \end{aligned}
    \end{equation}
    where $\Delta$, $B^d$, and $C^d$ are learnable linear functions of $x_t$. $\Delta_t$ represents the time step of the discretization. The selective mechanism in the Mamba block is commonly referred to as a selective scan. For better distinguishing with our 2D method, we refer to this scan as \textit{1D selective scan} due to its 1D scanning process. 
    \subsection{Architecture of \methodnameMIL~} 
    The overall architecture of \methodnameMIL~is illustrated in Fig.\ref{fig:2dmamba}. The model includes: $U$ layers of \methodname~blocks and an aggregator. The first component is the stack of $U$ \methodname~layers. We utilize the design of the original Mamba block \cite{mamba} and replace the original 1D selective scan with our 2D variant. The second component is an aggregator, which is an attention based module with two linear projections, producing a slide feature.\\
    \indent Our \methodnameMIL~first tiles an input WSI into patches $\{X_{i, j}\}$ with $i\in\{1,2,\dots, H\}$ and $j\in\{1,2,\dots, W\}$. $H$ and $W$ denote the number of the tiled patches along the height and width dimensions, respectively. These patches are then embedded differently based on their types. Tissue patches are embedded using a pre-trained pathology feature extractor $f$. Additionally, we propose using a learnable token $p$ to represent non-tissue patches padded to obtain a 2D rectangle feature map.
    This allows the model to learn the proper representation of non-tissue regions during training. Formally, the WSI is transformed into a feature map $x$ with a rectangle shape $(H, W)$: 
    \begin{align}
        x_{i, j} = 
        \begin{cases}
          f(X_{i, j}) & \text{if $X_{i, j}$ is a tissue patch}\\
          p & \text{otherwise}
        \end{cases}
        \ .
    \end{align}
    \subsection{2D selective SSM architecture.} We detail 2D selective SSM architecture. The key component of \methodname~is the \textit{2D selective scan} operation. In contrast to vanilla mamba which aggregates information from a flattened 1D sequence, \methodname~aggregates both geometric and semantic information directly from a 2D feature map. Particularly, \methodname~conducts both horizontal and vertical scans in parallel. For simplicity, we omit the state dimension superscript $d$ in this section. The parameters of the 2D selective scan remain the same as 1D in Eq.~\eqref{equ:ssm}, with the subscript being $(i, j)$ to index 2D inputs instead of $t$. We reuse $x_{i,j}$ to represent the input of the 2D selective scan after normalization, projection, and convolution layers in Fig.~\ref{fig:2dmamba}.\\
    \indent We formulate 2D scanning in a manner similar to the vanilla Mamba to maintain efficient parallelism. As shown in Fig.~\ref{fig:2dmamba}, we first conduct a horizontal scan on each row independently, equivalent to applying a 1D selective scan to each row. Specifically, the state $h_{i,j}^{\mathrm{hor}}$ obtained during the horizontal scan is:
    \begin{align}
        h_{i,j}^{\mathrm{hor}} &= \bar{A}_{i, j} h_{i, j-1}^{\mathrm{hor}} +  \bar{B}_{i, j} x_{i,j} 
        \ .
        \label{equ:h_scan}
    \end{align}
    Note that, for the first column, we assume $h_{i, 0}^{\mathrm{hor}} = 0$, and thus $h_{i,1}^{\mathrm{hor}} = \bar{B}_{i, 1} x_{i,1}$. Two parameters $\bar{A}_{i, j}$ and $\bar{B}_{i, j}$, which depend on $x_{i,j}$, regulates the information of previous states $h_{i, j-1}$ and current input $x_{i,j}$.\\
    \indent After the horizontal scan, we apply our vertical scan on each column of $h_{i,j}^{\mathrm{hor}}$ independently. Compared with horizontal scan,  we replace $\bar{B}_{i, j} x_{i,j}$ with the result $h_{i,j}^{\mathrm{hor}}$ obtained from the horizontal scan.
    \begin{align}
        h_{i,j} &= \bar{A}_{i, j} h_{i-1, j} + h_{i,j}^{\mathrm{hor}} \label{equ:v_scan}
        \ .
    \end{align}
    Note that for the first row $h_{1,j} = h^h_{1,j}$ by assuming $h_{0, j}^{\mathrm{hor}} = 0$. We reuse the same $\bar{A}_{i, j}$ for the vertical scan.\\
    \indent If we omit the subscripts of $\bar{A}$ and $\bar{B}$, and expand Eqs.~\eqref{equ:h_scan}~\eqref{equ:v_scan}, the hidden state $h_{i,j}$ can be formulated as the following equation (detailed derivations in Supp. \ref{derivation}):
    \begin{align}
        h_{i,j} = \sum_{i'\leq i} \sum_{j' \leq j} \bar{A}^{(i - i' + j - j')} \bar{B}x_{i',j'}\ ,
        \label{equ:2d_recursion}
    \end{align}
    After two scans, the output $y$ is aggregated from $h$ by the parameter $C$ similar to 1D-Mamba: $y_{i, j} = {C}h_{i, j}$. For each location $(i, j)$, the aggregation information is obtained from its upper left locations.\\
    \indent By doing so, \methodname~aggregates information without \textit{spatial discrepancy}. In comparison, the hidden state of the vanilla Mamba on a flattened image is given by $h^{1D}_{i} = \sum_{i'\leq i}\bar{A}^{\bm{i-i'}}\bar{B}x_{i'}$ where $i$ denotes the 1D index. The order $i$-$i'$ represents the distance between $i, i'$ in a flattened sequence, where a higher order (larger distance) may lead to forgetting \cite{shi2024multi}. This mathematically encapsulates the concept of ``\textit{spatial discrepancy}''. In contrast, \methodname~achieves the formulation in Eq.(\ref{equ:2d_recursion}), where the order $i$-$i'$+$j$-$j'$ corresponds to the Manhattan distance between $(i', j')$ and $(i, j)$, thereby preserving the 2D structure. This distance represents a path from $(i', j')$ to $(i, j)$ that moves horizontally to the right and then vertically downward. This mathematically encapsulates the concept of ``\textit{spatial continuity}''. For instance, the last hidden state for a 3x3 feature map can be expressed as:
    \begin{align}
        \resizebox{0.90\hsize}{!}{%
        \begin{math}
        \begin{aligned}
        h^{1D}_{3, 3} = &\bar{A}^{\cbm{8}}\bar{B}x_{1,1} + \bar{A}^{\cbm{7}}\bar{B}x_{1,2} + \bar{A}^{\cbm{6}}\bar{B}x_{1,3} + & h^{2D}_{3, 3} =& \bar{A}^{\cbm{4}}\bar{B}x_{1,1} + \bar{A}^{\cbm{3}}\bar{B}x_{1,2} + \bar{A}^{\cbm{2}}\bar{B}x_{1,3} +\\
        &\bar{A}^{\cbm{5}}\bar{B}x_{2,1} + \bar{A}^{\cbm{4}}\bar{B}x_{2,2} + \bar{A}^{\cbm{3}}\bar{B}x_{2,3} + & &  \bar{A}^{\cbm{3}}\bar{B}x_{2,1} + \bar{A}^{\cbm{2}}\bar{B}x_{2,2} + \bar{A}^{\cbm{1}}\bar{B}x_{2,3} +\\
        &\bar{A}^{\cbm{2}}\bar{B}x_{3,1} + \bar{A}^{\cbm{1}}\bar{B}x_{3,2} + \bar{A}^{\cbm{0}}\bar{B}x_{3,3}  & & \bar{A}^{\cbm{2}}\bar{B}x_{3,1} + \bar{A}^{\cbm{1}}\bar{B}x_{3,2} + \bar{A}^{\cbm{0}}\bar{B}x_{3,3}
        \end{aligned}
        \end{math}%
    }
    \end{align}
    \noindent where a much larger order term $\bar{A}^8$ of $x_{1,1}$ in the 1D case (compared to $\bar{A}^4$ in 2D case) results in much more forgetting and a loss of the 2D structure information. Notably, \textit{spatial discrepancy} becomes particularly problematic in WSIs,  as it leads to significantly larger order terms (e.g. $\bar{A}^{200}$) due to the large size of WSIs.
\begin{figure*}[ht] 
    \centering
    \includegraphics[width=1\linewidth]{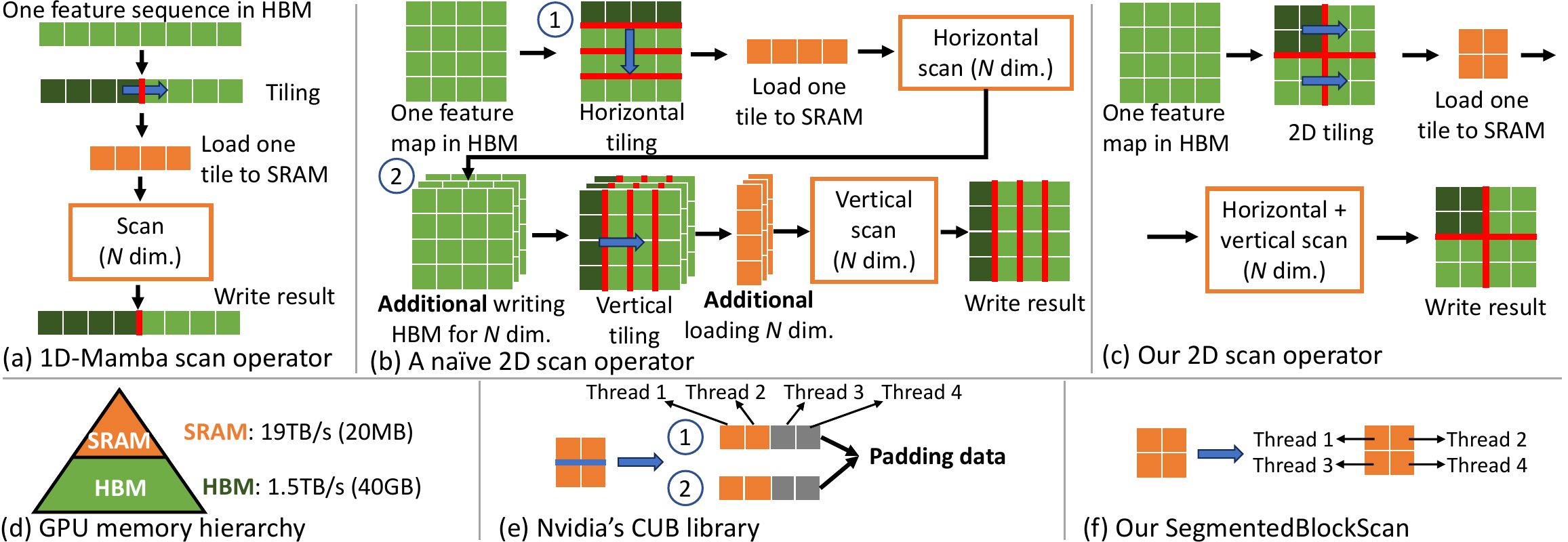}
    \caption{Our hardware-aware 2D selective scan operator with efficient caching mechanism and high parallelism. {\color{myorange} {Orange}} color represents operations on SRAM and {\color{mygreen} {green}} color represents those on HBM. (a) The \textbf{1D Mamba scan operator} intakes a flattened sequence on HBM. It tiles the input into sub-sequences. Each sub-sequence is loaded from HBM to SRAM, scanned and reduced across $N$ intermediate dimensions, and then written back to HBM. The total memory access complexity is $\mathcal{O}(L)$. (b) A \textbf{Naive 2D scan operator} tiles the 2D feature map by rows and columns, and performs 1D Mamba scans on each row, column, and on $N$ independent state dimensions. This will explicitly instantiate $N$ intermediate feature maps on HBM, resulting in a memory access complexity of $\mathcal{O}(NL)$. (c) \textbf{Our 2D scan operator} tiles the feature map into 2D grids and scans each grid in 2 directions. Intermediate features are reduced inside each tile, only the aggregated result is stored back to HBM. The memory complexity is $\mathcal{O}(L)$. (d) \textbf{GPU memory hierarchy}: SRAMs are small but fast; HBMs have large capacities but are slow. (e) \textbf{NVIDIA's CUB BlockScan} only supports 1D sequences, with sizes of multiples of $32$. Scanning a two-row grid requires two sequential kernel launches and padding elements. (f) \textbf{Our SegmentedBlockScan} enables scanning multiple rows and columns in parallel. It reduces the amount of memory transactions and padding data.}
    \label{fig:2dmamba_cuda}
\end{figure*}

\subsection{Hardware-Aware 2D Selective Scan}
\label{subsec:method_cuda}
    We present our \textit{hardware-aware scanning operator} that accelerates 2D selective scans. First, we revisit the GPU memory hierarchy and analyze the major challenges for 2D selective scans. Then, we present our novel operator in detail.
    
    \textbf{GPU memory hierarchy}. Fig.~\ref{fig:2dmamba_cuda} (d) illustrates the memory hierarchy of modern GPUs. The {\color{mygreen} {green}} area represents off-chip GPU memory, with low speed and high capacity. Here, it is referred to as \textit{high bandwidth memory} (HBM). The {\color{myorange} {orange}} area denotes on-chip memory, with high speed but low capacity, and is referred to as SRAM. In GPU algorithms, data is transferred from HBM to SRAM for computation, and the results are stored back to HBM to vacate SRAM for succeeding computation. Memory transfers are expensive. Therefore, instead of computation, many GPU algorithms \cite{dao2022flashattention,dao2023flashattention2} are bounded by memory. Mamba's selective scan \cite{mamba} is also \textit{memory-bounded}.
    
    \textbf{Mamba's 1D selective scan}. The vanilla Mamba is fast as the GPU memory follows a hierarchy by 1D tiling and caching. As shown in Fig.~\ref{fig:2dmamba_cuda} (a), a long  sequence in HBM is divided into smaller tiles. Each tile is loaded into SRAM, scanned across $N$ independent state dimensions, aggregated into a single output by rules specified in Eq.~\eqref{equ:mamba_y}, and stored back to HBM. The intermediate results of the $N$ state dimensions are \textit{\textbf{not} materialized on HBM}, and will be recomputed during back-propagation. The overall memory access complexity is $\mathcal{O}(L)$, where $L$ denotes the sequence length.
    
    \textbf{Naive 2D selective scan}. It is not trivial to extend 1D Mamba scans to 2D. As illustrated in Fig.~\ref{fig:2dmamba_cuda} (b), a naive extension to 1D Mamba will scan a 2D feature map in two steps. First, the feature map is tiled into $H$ rows for row-wise 1D Mamba scans. Next, the succeeding vertical scans must be applied independently to each column, where each column has $N$ independent state dimensions. Therefore, the horizontal scanner \textit{\textbf{must} materialize} $N$ intermediate feature maps on HBM. Each feature map is then tiled into $W$ columns for column-wise 1D Mamba scans. Its memory access complexity is $\mathcal{O}(NHW) = \mathcal{O}(NL)$, which, as demonstrated in Table~\ref{tab:th_mem}, results in low throughput and high memory consumption.
    
    \textbf{Hardware-aware 2D selective scan}. The proposed hardware-aware 2D selective scan operator, which is illustrated in Fig.~\ref{fig:2dmamba_cuda} (c), optimizes memory transactions by 2D tiling and caching. Instead of tiling by rows or by columns, we divide the feature map into a 2D grid. At each step, we only load a small submatrix into SRAM. Then we conduct horizontal and vertical scans for $N$ independent state dimensions, and write the aggregated output back to HBM. This \textit{\textbf{avoids} the explicit materialization} of the state dimensions, and maintains an overall memory access complexity of $\mathcal{O}(HW) = \mathcal{O}(L)$, equivalent to the vanilla Mamba.
    
    Moreover, vanilla Mamba employs NVIDIA's CUB library~\cite{nvidia24-online-cub} for 1D parallel scans. However, as illustrated in Fig.~\ref{fig:2dmamba_cuda} (e), CUB's BlockScan algorithm only supports full-sequence scanning. Thus, it requires multiple scans for a multi-row feature map. Moreover, for a 2D feature map, CUB BlockScan requires both its height $H$ and width $W$ to be multiples of $32$, where $32$ is the smallest thread schedule granularity for NVIDIA GPUs. Therefore, small feature maps must be padded before computation, leading to inefficiencies. For instance, a typical $14 \times 14$ feature map will require $18$ padding elements per row and column, wasting as much as $56\%$ of computation. To resolve this limitation, we introduce the SegmentedBlockScan algorithm, which is illustrated in Fig.~\ref{fig:2dmamba_cuda} (f). It distributes GPU threads across both rows and columns, only requiring $H \times W$ to be a multiple of $32$. This enables simultaneous multi-row/column scanning and significantly reduces the padding requirements for small feature maps. For instance, regarding the same $14 \times 14$ feature map, our method requires only $2$ padding elements per row and column. Detailed algorithm and implementations can be found in Supp. \ref{sec:supp:cuda}.
\begin{table*}[ht]
    \centering
    \resizebox{1\textwidth}{!}{
    \setlength{\tabcolsep}{0.9mm}{
    \begin{tabular} {l ccc c ccc c ccc c ccc c ccc}
        \toprule
        \multirow{2}{*}{\textbf{Method}} & \multicolumn{3}{c}{\textbf{BRACS}} && \multicolumn{3}{c}{\textbf{DHMC}} && \multicolumn{3}{c}{\textbf{PANDA}} && \multicolumn{3}{c}{\textbf{TCGA-NSCLC}}  && \multicolumn{3}{c}{\textbf{TCGA-BRCA}}\\ 
        \cline{2-4}  \cline{6-8} \cline{10-12} \cline{14-16} \cline{18-20}& 
        $Acc$ & $F1$  & $AUC$ && 
        $Acc$ & $F1$  & $AUC$ &&
        $Acc$ & $F1$  & $AUC$ &&
        $Acc$ & $F1$  & $AUC$ &&
        $Acc$ & $F1$  & $AUC$ \\            
        \midrule
        AB-MIL  & 
        0.7057  & 0.6015 & 0.8939 &&  
        0.8684  & 0.7774 & {0.9695} && 
        0.4883  & 0.4269 & 0.7797 &&  
        0.8758  & 0.8756 & 0.9572 &&
        0.9292  & 0.8893 & 0.9747\\ 
        DSMIL & 
        0.6759  & 0.5618 & 0.8618 &&  
        {0.8711}  & {0.7934} & 0.9583 &&
        0.4633  & 0.3847 & 0.7660 &&  
        0.8782  & 0.8780 & 0.9567 &&
        0.9375  & 0.8961 & 0.9770\\ 
        CLAM & 
        0.7103  & 0.6014 & \textbf{0.9016} &&  
        {0.8711}  & 0.7909 & \textbf{0.9727} && 
        0.4802  & 0.4224 & 0.7820 &&  
        0.8804  & 0.8803 & 0.9536 &&
        0.9333  & 0.8960 & 0.9753\\ 
        DTFD-MIL  & 
        0.7012  & 0.6131 & 0.8787 &&  
        {0.8711}  & 0.7704 & 0.9521 &&
        0.4704  & 0.3853 & 0.7665 &&
        0.8736  & 0.8732 & 0.9559 &&
        0.9271  & 0.8809 & 0.9633\\
        TransMIL  & 
        0.6919  & 0.6063 & 0.8759  &&  
        0.8067  & 0.7136 & 0.9466  && 
        0.4636  & 0.3970 & 0.7728  &&  
        {0.8850}  & 0.8845 & \textbf{0.9626}&&
        0.9375  & 0.9028 & 0.9763\\  
        \midrule
        S4-MIL & 
        0.6621  & 0.5904 & 0.8457 &&  
        0.8644  & 0.7847 & 0.9284 && 
        {0.5047}  & {0.4486} & {0.7986} &&  
        \textbf{0.8851} & {0.8849} & 0.9571&&
        \textbf{0.9458}  & 0.9154 & 0.9770\\
        MambaMIL & 
        {0.7379}  & {0.6832} & 0.8883 &&  
        0.8550  & 0.7789 & 0.9661 && 
        0.4679  & 0.4216 & 0.7781 &&  
        0.8758  & 0.8756 & 0.9582 &&
        0.9333  & 0.8939 & 0.9657\\
        SRMambaMIL &  
        {0.7379}  & 0.6789 & 0.8915 &&  
        0.8590  & 0.7735 & 0.9639 && 
        0.4711  & 0.4209 & 0.7776  &&  
        {0.8850}  & {0.8849} & 0.9592 &&
        0.9313  & 0.8900 & 0.9657\\
        \midrule
        \textbf{\methodnameMIL}& 
        \textbf{0.7517} & \textbf{0.7045} & {0.8964} &&  
        \textbf{0.8926} & \textbf{0.8027} & 0.9468 && 
        \textbf{0.5075} & \textbf{0.4562} & \textbf{0.8184}  &&  
        \textbf{0.8851} & \textbf{0.8850} & {0.9618} &&
        \textbf{0.9458} & \textbf{0.9156} & \textbf{0.9782}\\ 
        \bottomrule
    \end{tabular}
    }
    }
    \caption{The comparison of accuracy (Acc), F1 and AUC on five \textbf{WSI classification} datasets. We conducted each experiment five times using five different random seeds and reported their mean. The highest metrics are marked as \textbf{bold}.}
    \label{tab:result_classification}
\end{table*}
\section{Experiments}
\subsection{Dataset}
We assess \methodnameMIL~on 5 public pathology classification datasets, TCGA-BRCA \cite{tcga}, BRACS \cite{bracs}, PANDA \cite{panda}, DHMC \cite{dhmc}, TCGA-NSCLC and 5 public survival datasets, TCGA-(KIRC, KIRP, LUAD, STAD, UCEC). These datasets cover a variety of organs including breast, prostate, lung, kidney, stomach, and uterine. The number of slides ranges from $261$ to $10614$ and we use $20$x magnification for all these datasets. Details of the datasets are listed in the Supp. \ref{sec:supp:dataset}. Following \cite{vmamba}, we evaluate our \methodname~on two natural image datasets of ImageNet-1K classification and ADE20K semantic segmentation.
\subsection{Results}
\textbf{WSI Classification.} We compare \methodnameMIL~ with eight other SOTA MILs on five WSI classification datasets. The baselines include ABMIL \cite{abmil}, CLAM \cite{clam}, DSMIL \cite{dsmil}, DTFDMIL \cite{zhang2022dtfd}, TransMIL \cite{shao2021transmil}, S4-MIL \cite{s4mil}, MambaMIL \cite{mambamil} and SRMambaMIL \cite{mambamil}. The first four MILs are attention-based, TransMIL is Transformer-based, and the last three are 1D SSM-based MILs. We utilize three metrics to evaluate the WSI classification performance: accuracy (\textit{Acc}), F1 score (\textit{F1}), and area under the curve (\textit{AUC}). Table \ref{tab:result_classification} shows that our \methodnameMIL~surpasses all current SOTA methods across multiple datasets, indicating our strong generalization ability. Compared with the best-performing non-Mamba method, we achieve significant improvements of up to $5.83\%$ in accuracy, $14.90\%$ in F1 score, and $4.65\%$ in AUC. \methodnameMIL~also outperforms SSM-based methods by up to $3.26\%$ in accuracy, $3.11\%$ in F1 score, and $2.48\%$ in AUC, showing the benefit of preserving spatial continuity in WSIs.\\
\indent\textbf{WSI Survival Analysis.} We further compare \methodnameMIL~with all eight MILs on five WSI survival datasets. We assess the performance using the concordance index (C-index), which evaluates how well a survival model ranks patients with their survival time compared to the actual survival outcomes. As shown in Table \ref{tab:result_survival_prediction}, \methodnameMIL~consistently achieves the highest C-index scores across all datasets, indicating superior predictive performance. Specifically, \methodnameMIL~achieves a relative improvement of $0.6\%$, $1.2\%$, $5.5\%$, $2.9\%,$ and $1.0\%$ on C-index compared with the best-performing baseline on KIRC, KIRP, LUAD, STAD, and UCEC, respectively.
\begin{table}[h]
\centering
\resizebox{0.47\textwidth}{!}{
\setlength{\tabcolsep}{1.7mm}{
\begin{tabular} {l c c c c c}
    \toprule
    \multirow{1}{*}{\textbf{Method}} & \multicolumn{1}{c}{\textbf{KIRC}} & \multicolumn{1}{c}{\textbf{KIRP}} & \multicolumn{1}{c}{\textbf{LUAD}} & \multicolumn{1}{c}{\textbf{STAD}} & \multicolumn{1}{c}{\textbf{UCEC}} \\
    \hline
    ABMIL  & 
    0.7051  &  0.7824  & 0.6157 &  0.6119  & 0.7243 \\ 
    DSMIL & 
    0.6240  &  0.7122  & 0.6114 &  0.6010  & 0.6324\\ 
    CLAM & 
    0.5723  &  0.7197  & 0.5874 &  0.5883  & 0.6312 \\
    DTFD-MIL  & 
    0.7271  &  0.7933  & 0.6020 &  0.6168  & 0.7462 \\
    TransMIL  & 
    0.6944  &  0.7317  & 0.6139 &  0.5978  & 0.6997 \\ \midrule
    S4-MIL & 
    0.7232  &  0.7905  & 0.5945 &  0.6001  & 0.7459 \\ 
    MambaMIL & 
    0.7096  &  0.7822  & 0.5952 &  0.6244  & 0.7419 \\ 
    SRMambaMIL &  
    0.7178  &  0.7424  & 0.5876 &  0.6130  & 0.7398 \\ 
    \midrule
    \textbf{\methodnameMIL~}& 
    \textbf{0.7311}  &  \textbf{0.8027}  & \textbf{0.6198} &  \textbf{0.6428}  & \textbf{0.7536} \\ 
    \bottomrule
\end{tabular}
}
}
\caption{The comparison of C-Index on five \textbf{survival analysis} datasets. We performed 5-fold cross-validation for all experiments. The highest metrics are \textbf{bold}.}
\label{tab:result_survival_prediction}
\end{table}

\textbf{Speed and GPU Memory Efficiency.} Our method demonstrates high speed and less memory usage. We evaluate the floating-point operations (FLOPs), throughput, and GPU memory consumption in inference on three input feature sizes: $14\times14$, $56\times56$, and $200\times200$. First, we compare three CUDA-based scanning operators: the CUB 1D scan used by Mamba, the naive 2D scan introduced in Section \ref{subsec:method_cuda}, and our optimized 2D scan, across the three input sizes with 16 independent state dimensions. As shown in Table \ref{tab:th_mem}, our 2D scan significantly outperforms the naive 2D scan in both throughput and GPU memory efficiency across all input sizes, with the performance gap widening as the feature size increases. Our 2D scan matches the throughput of Mamba's CUB scan for the $14\times14$ input size. However, as input size increases, its throughput declines compared to the CUB scan. This is due to more complex memory layout of 2D data and our doubled computations. Nonetheless, our 2D scan maintains linear memory consumption with respect to the sequence length. We then assess the CUDA implementation of Mamba, the Python implementation of our \methodname, and the CUDA implementation of our \methodname~within the MIL framework, across the three input feature sizes. Table \ref{tab:th_mem} indicates that our CUDA-based \methodnameMIL~framework consistently outperforms the Python-based implementation across all metrics, benefiting from our hardware-aware 2D scan operator. The throughput of our method remains at $70\%$-$90\%$ of the vanilla Mamba-based MIL framework.
\begin{table*}[h]
    \centering
    \small
    \setlength{\tabcolsep}{1.1mm}{
    \begin{tabular}{llccc@{\extracolsep{3pt}}ccc@{\extracolsep{3pt}}ccc}
    \toprule
    &Feature size                            &
    \multicolumn{3}{c}{$14\times14$} & \multicolumn{3}{c}{$56\times56$} & \multicolumn{3}{c}{$200\times200$} \\
    \cline{3-5} \cline{6-8} \cline{9-11}
    Scope & Method                                   & FLOPs    & Thro.    & GPU Mem.   & FLOPs    & Thro.    & GPU Mem.   & FLOPs     & Thro.    & GPU Mem.    \\ \midrule
                    & CUB 1D scan                    & 9K     & 49K        & 1.6KB      & 150K    & 12K       & 25.1KB     & 1.9M      & 3K        & 0.3MB       \\
    CUDA operator   & Naive 2D scan                  & 16K    & 0.2K       & 14.1KB     & 251K    & 0.06K     & 225.8KB    & 3.2M      & 0.02K     & 2.9MB       \\
                    & \textbf{Our 2D scan}           & 16K    & 40K        & 1.6KB      & 251K    & 6K        & 38.5KB     & 3.2M      & 1K        & 0.5MB        \\ \midrule
                    & Mamba (CUDA)                    & 58M      & 894      & 24MB      & 0.9G     & 752      & 58MB      & 11.8G     & 203      & 500MB      \\
    MIL framework   & \methodname~(Python)            & 63M      & 327      & 46MB      & 1.0G     & 187      & 430MB     & 12.9G     & 13       & 5842MB     \\
                    & \textbf{\methodname~(CUDA)}     & 63M      & 655      & 24MB      & 1.0G     & 625      & 76MB     & 12.9G     & 185      & 598MB      \\
    \bottomrule
    \end{tabular}
    }
    \caption{Comparison of floating-point operations (FLOPs), throughput (Thro., feature maps per second), and GPU memory consumption during inference. CUDA operators are measured using single dimensional feature input and MIL frameworks are measured using $128$ dimensional feature input. The state dimension is set to $16$ for all experiments.}
    \label{tab:th_mem}
\end{table*}

\textbf{Natural Image Classification.} Besides its effectiveness and efficiency on pathology images, our method also generalizes well on natural image classifications. We apply our \methodname~to the SOTA Mamba-based method on natural images, VMamba~\cite{vmamba}. We replace its Mamba block with our \methodname~block and name it as 2DVMamba. We first evaluate it on the ImageNet-1K classification dataset and compare them with: Swin Transformer~\cite{liu2021Swin}, Vim~\cite{visionmamba}, EfficientVMamba~\cite{pei2024efficientvmamba}, LocalVMamba~\cite{localmamba} and the original VMamba. Table \ref{tab:imagenet1k} shows that our 2DVMamba achieves $0.2\%$ higher accuracy than the original VMamba and surpasses all SOTA methods.

\begin{table}[h]
\centering
\small
\begin{tabular}{lccc}
\toprule
Method                  & \#Param & FLOPs & Top-1 Acc\%   \\
\midrule
Swin-T                  & 28M     &  4.5G & 81.3          \\
Vim-S                   & 26M     &   -   & 80.3          \\
EfficientVMamba-B       & 33M     &  4.0G & 81.8          \\
LocalVMamba-T           & 26M     &  5.7G & 82.7          \\
VMamba-T                & 30M     &  4.91G & 82.6          \\
\midrule
\textbf{2DVMamba-T}     & 30M     &  4.94G & \textbf{82.8} \\
\bottomrule
\end{tabular}

\caption{The top-1 accuracy (\%) of our V2DMamba-T on the ImageNet-1K dataset. All images are of size $224 \times 224$.}
\label{tab:imagenet1k}
\end{table}

\textbf{Natural Image Segmentation.} We further evaluate the performance of 2DVMamba on the ADE20K semantic segmentation dataset. Tab. \ref{tab:ade20k} shows that 2DVMamba-T outperforms the baseline VMamba-T, achieving a gain of $0.7$ in single-scale mIoU, $0.5$ in multi-scale mIoU and surpassing all baselines. Notably, the enhancement in segmentation performance is more pronounced compared to that in classification. This is likely due to that segmentation is a dense prediction task where maintaining spatial continuity across patches is crucial.

\begin{table}[h]
\centering
\small
\setlength{\tabcolsep}{0.9mm}{
\begin{tabular}{lcccc}
\toprule
Method                  & \#Param. & FLOPs & mIoU(SS)      & mIoU(MS)      \\
\midrule
Swin-T                  &   60M    &  945G & 44.5          & 45.8          \\
Vim-S                   &   46M    &   -   & 44.9          & -             \\
EfficientVMamba-B       &   65M    &  930G & 46.5          & 47.3          \\
LocalVMamba-T           &   57M    &  970G & 47.9          & 49.1          \\
VMamba-T                &   62M    &  949G & 47.9          & 48.8          \\
\midrule
\textbf{2DVMamba-T}     &   62M    &  950G & \textbf{48.6} & \textbf{49.3} \\
\bottomrule
\end{tabular}
}
\caption{The performance of our 2DVMamba-T on the ADE20K semantic segmentation dataset. ``SS'' and ``MS'' denote single-scale and multi-scale testing, respectively. FLOPs are calculated with an input size of $512\times2048$.}
\label{tab:ade20k}
\end{table}

\begin{figure*}[t] 
      \centering
      \includegraphics[width=0.97\linewidth]{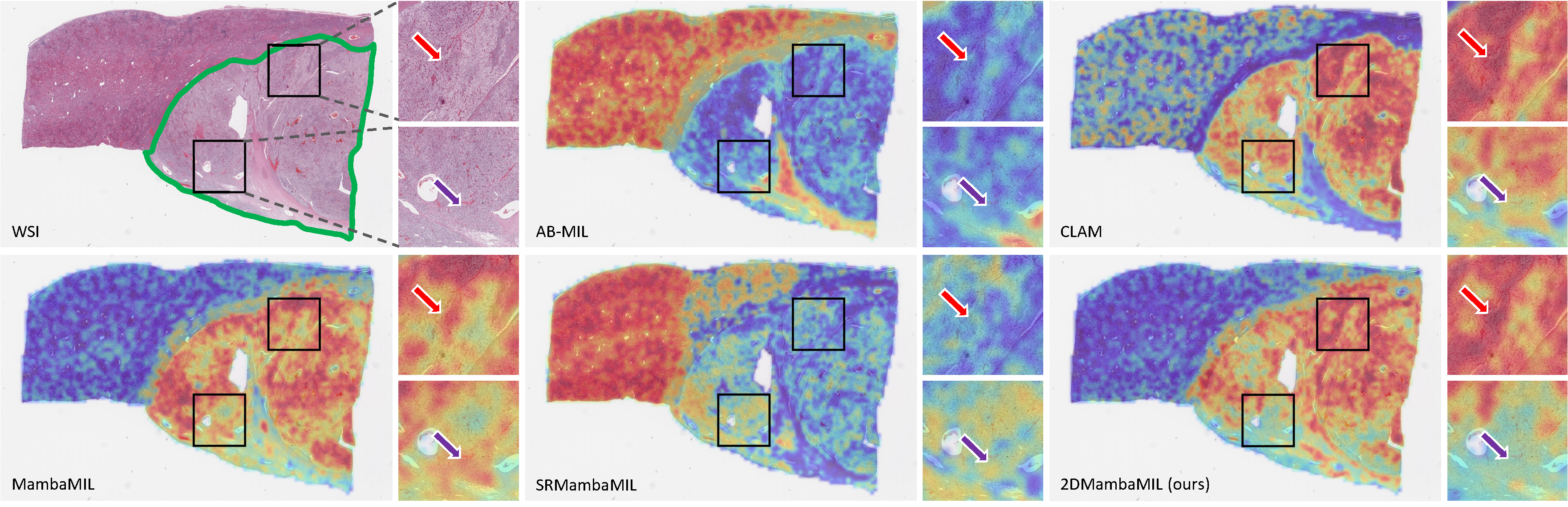}
      \caption{
      The attention visualization of \methodnameMIL~and four other methods on a TCGA-KIRC sample for survival analysis. Tumor regions are outlined in {\color{mygreen}green}. AB-MIL and SRMambaMIL primarily focus on non-tumor areas, while CLAM also shows substantial attention to non-tumor regions. In contrast, both \methodnameMIL~and MambaMIL focus predominantly on tumor regions. Compared with MambaMIL, attention of \methodnameMIL~shows a more heterogeneous fashion, focusing more on critical regions related to survival ({\color{red}red} arrows) while paying less attention to less related ones ({\color{violet}violet} arrows).}
      \label{fig:heatmap}
\end{figure*}

\textbf{Ablation on the non-tissue padding.} We ablate our learnable padding token for non-tissue regions by comparing it with a naive solution: padding all fixed zero tokens, on the PANDA and TCGA-BRCA datasets. Table \ref{tab:ab_padding_token} shows that our learnable padding outperforms the fixed padding by relatively $1.56\%$-$4.25\%$ and $0.62\%$-$1.58\%$ in accuracy and AUC, respectively. This suggests that our trainable padding  enables the scanning to adapt more effectively to the non-tissue regions. 

    \begin{table}[h]
        \centering
        \small
        \begin{tabular} {c cc cc}
            \toprule
            \multirow{2}{*}{\textbf{Padding token}} &  \multicolumn{2}{c}{\textbf{PANDA}} & \multicolumn{2}{c}{\textbf{TCGA-BRCA}} \\ 
            \cline{2-5}  & 
            $Acc$ & $AUC$ & 
            $Acc$ & $AUC$ \\         
            \hline
            Fixed zero & 
            0.4868  & 0.8057 &  
            0.9313  & 0.9722 \\
            \textbf{Learnable} & 
            \textbf{0.5075} & \textbf{0.8184}&  
            \textbf{0.9458} & \textbf{0.9782} \\ 
            \bottomrule
        \end{tabular}
        \caption{Ablation on the non-tissue paddings on the PANDA and TCGA-BRCA dataset. Our learnable token achieves higher performance compared to the fixed zero token.}
        \label{tab:ab_padding_token}
    \end{table}
    \textbf{Ablation on the multi-directional scanning}. We ablate MambaMIL using 2-direction \cite{visionmamba}, 4-direction raster \cite{plainmamba}, and 4-direction cross scans \cite{vmamba}, comparing with 2DMambaMIL. Tab.\ref{tab:1dvs2d} shows that while scanning in multiple directions improves performance, it remains inferior to 2DMambaMIL. This demonstrates that multi-directional scanning does not accurately model cell-to-cell interactions, as cells interact in a coordinated manner across all directions, rather than being limited to horizontal and vertical orientations.
    \begin{table}[h]
    \centering
    \small
    \setlength{\tabcolsep}{2.0mm}{
    \begin{tabular} {l cc cc}
        \toprule
        \multirow{2}{*}{\textbf{Method}} &  \multicolumn{2}{c}{\textbf{PANDA}} & \multicolumn{2}{c}{\textbf{TCGA-BRCA}} \\ 
        \cline{2-5}  & 
        $Acc$ & $AUC$ & 
        $Acc$ & $AUC$ \\         
        \hline
         MambaMIL (1D) & 
         0.4679 & 0.7781 & 
         0.9333 & 0.9657 \\
         w. 2-direction & 
         0.4853  & 0.7749 &
         0.9374  & 0.9753 \\
         w. 4-direction (raster)& 
         0.4923 &  0.7918 &
         0.9388  & 0.9755 \\
         w. 4-direction (cross)& 
         0.4939 & 0.8006 &
         0.9402  & 0.9698 \\         
         \midrule
        \textbf{\methodnameMIL~(2D)} & 
        \textbf{0.5075} & \textbf{0.8184} &  
        \textbf{0.9458} & \textbf{0.9782} \\
        \bottomrule
    \end{tabular}
    }
    \caption{The comparison of MambaMIL with 2-direction \cite{visionmamba}, 4-direction scan \cite{plainmamba,vmamba} and 2DMambaMIL. }
    \label{tab:1dvs2d}
    \end{table}

\textbf{Qualitative Evaluation.} We qualitatively compare the attention heatmaps generated by \methodnameMIL~with four existing approaches (AB-MIL, CLAM, MambaMIL, and SRMambaMIL) for classification and survival analysis tasks, focusing on pathological and biological interpretability. The results demonstrate that \methodnameMIL~consistently targets tumor areas in WSI classification and survival analysis datasets, occasionally including pixels from the immediate tumor-adjacent regions. Fig.~\ref{fig:heatmap} presents a case of kidney clear cell carcinoma in the context of survival analysis. AB-MIL and SRMambaMIL predominantly attend to non-tumor regions, which is unnecessary for risk prediction, and CLAM also shows considerable attention to non-tumor areas. On the contrary, the attentions of our \methodnameMIL~ and SRMambaMIL are both driven by tumor areas. In comparison, our method exhibits a more heterogeneous attention pattern, specifically focusing on highly survival-related regions (indicated by the {\color{red}red} arrows), whereas SRMambaMIL's attention is more uniformly distributed, which focuses on some less survival-related regions (indicated by the {\color{violet}violet} arrows). Additional qualitative evaluations are shown in Supp. \ref{sec:supp:qualitative}.

\begin{figure}[htp] 
      \centering
      \includegraphics[width=0.75\linewidth]{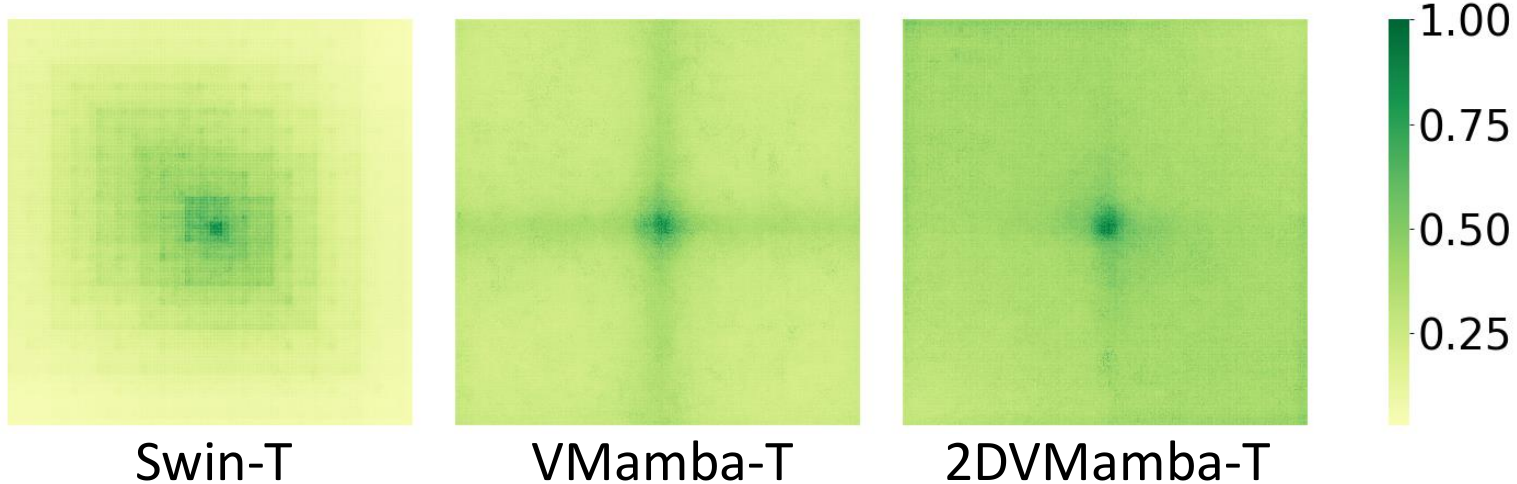}
      \caption{Comparison of Effective Receptive Fields (ERF) between Swin-T, VMamba-T and 2DVMamba-T. Pixels with higher intensity indicate larger responses regarding the central pixel.}
      \label{fig:erf}
\end{figure}

\textbf{Visualization of Effective Receptive Fields.} Effective Receptive Fields (ERF) refers to the region in the input space that contributes to the activation of a certain output unit \cite{luo2016understanding}. We conduct analyses of the central pixel’s ERF between Swin-T, VMamaba-T and 2DVMamba-T. Fig.\ref{fig:erf} shows that the ERF of Swin-T expresses a local pattern, consistent with its local structure. VMamaba-T expresses a more global pattern but  has a clear cross signal resulting from its 4-way 1D scan process. 2DVMamba demonstrates much more global and smooth ERFs without cross-signal showcasing to persevere spatial continuity.

\section{Conclusion}
\label{sec:conclusion}
In this work, we presented \methodname, a novel 2D selective SSM framework incorporating the 2D spatial structure of images into the Mamba paradigm. Unlike the vanilla Mamba processing a flattened 1D sequence, \methodname~employs a 2D selective SSM architecture to capture both geometric and semantic information directly from 2D feature maps using a hardware-optimized selective scan operator. We evaluated \methodname~on 10 WSI classification and survival analysis datasets, where our method consistently outperformed conventional MIL methods and 1D Mamba-based MIL approaches. Furthermore, our design enhances GPU efficiency through optimized caching and increased parallelism for 2D structures. We also demonstrated its strong generalizability on natural image classification and segmentation tasks by integrating it into a SOTA method. Future work will focus on refining 2D SSM designs and exploring broader applications.

\section*{Acknowledgement} This work was partially supported by USA NSF grants IIS-2123920 [D.S], IIS-2212046 [D.S], IIS-1715985 [H.Q], IIS-1812606 [H.Q], the Canadian Cancer Society Breakthrough Grant [V.Q.H.T], FRQS-CRS-J1 [V.Q.H.T], NSERC-DG RGPIN‐2022‐05378 [M.S.H] and Amazon Research Award [M.S.H]. The computational pathology experiments reported in this paper were executed on Korea University servers administered by Professor Jin Tae Kwak.

{
    \small
    \bibliographystyle{ieeenat_fullname}
    \bibliography{main}
}

\clearpage
\setcounter{page}{1}
\setcounter{section}{0}
\maketitlesupplementary
\beginsupplement
\renewcommand{\thesection}{\Alph{section}}

\section{Implementation Details}
\noindent\textbf{SSM.} For a fair comparison, we use a single SSM-based block with a $128$-dimensional SSM and set the state dimension to $16$ for all Mamba-based methods.

\noindent\textbf{Feature extractor.} We use UNI \cite{chen2024uni}, a well-known and current SOTA foundation model for feature extraction, which is a ViT-L/16 pretrained on more than 100 million pathology patches from from over 100,000 H\&E-stained WSIs across 20 major tissue types. UNI is pretrained in a self-supervised manner using DINOv2 \cite{oquab2023dinov2}. 

\noindent\textbf{Aggregator.} The aggregator in our \methodname~follows \cite{gu2021combining,transformer} using attention pooling \cite{gu2021combining} and two linear layers (128 intermediate dimensions). This module produce the attention scores of each patch embedding, then aggregate them using the weighted summation to produce the WSI embedding. Mathematically, let $H=\{{h}_{1}, \ldots , {h}_{K}\}$ be a bag of $K$ embeddings, the attention pooling is:
\begin{equation}\label{eq:weighted_sum}
    {z} = \sum_{k=1}^{K} a_k {h}_k,
\end{equation}
where:
\begin{equation}\label{eq:attention}
a_{k} = \dfrac{ \exp\{{w}^{\top} \tanh \big{(} {V} {h}_{k}^{\top} \big{)}\}}{{\displaystyle \sum_{j=1}^K} \exp \{ {w}^{\top} \tanh \big{(}{V} {h}_{j}^{\top}\big{)} \} },
\end{equation}
where ${w} \in \mathbb{R}^{128\times 1}$ and ${V} \in \mathbb{R}^{128 \times M}$ are parameters ($M$ is the embedding dimension). For fair comparisons, we use the same feature extractor and aggregator as other MILs.

\noindent\textbf{WSI pre-processing.} We extract patches from WSIs at 20x magnification with no overlapping. The patch size is set to 512x512 pixels. We used the preprocessing tool in CLAM \cite{clam} to segment and extract tissue regions.

\noindent\textbf{Training.} We use AdamW \cite{adamw} to optimize the models for $20$ epochs of training with batch size being 1. The initial learning rate is set to $0.0001$ and is adjusted with a cosine annealing scheduler. All pathology and natural image experiments are trained using one NVIDIA V100 GPU and eight NVIDIA A100 GPUs, respectively.

\section{Details of the hardware-aware 2D selective scan operator}
\label{sec:supp:cuda}
In this section, we detail the hardware-aware 2D selective scan operator in 2DMamba introduced in section \ref{subsec:method_cuda}. The forward pass of the 2D selective scan, implemented as a fused kernel with 2D tiling, is formulated in the algorithm below. The backward pass of 2DMamba follows a similar structure but involves four 2D selective scans: one horizontal and one vertical scan to reconstruct intermediate variables from the forward pass, and one horizontal reverse and one vertical reverse scan to propagate gradients. 

\newcommand{\algorithmicpara}{\textbf{SSM parameters:}}
\newcommand{\Parameters}{\item[\algorithmicpara]}
\renewcommand{\thealgorithm}{}

\begin{algorithm}[h]
    \caption{2D selective scan (fused kernel with tiling)}
    \label{alg:2dmamba_forward}
    \small
    \begin{algorithmic}
        \Require{2D input feature $x:(H, W)$; Time step $\Delta:(H, W)$;}
        \Require{State dimension $N$. Tile size $T$.}
        \Parameters{Input independent $A:(N)$, $D$ and $bias$;}
        \Parameters{Input dependent $B:(H, W)$;}
        \Parameters{Input dependent $C:(H, W)$.}
        \OUTPUT{2D aggregated result $y$}.
        
        \State $K_H = \left\lceil H / T \right\rceil$
        \State $K_W = \left\lceil W / T \right\rceil$
        \State \textcolor{brown}{\textit{\# loop for $K_H * K_W$ tiles}}
        \For{$k_h = 1$ to $K_H$ and $k_w = 1$ to $K_W$}
            \State {$x_{k_h, k_w}:(T, T)$ = Read from HBM.}
            \State {$\Delta_{k_h, k_w}:(T,T)$ Read from HBM.}
            \State $\Delta_x = softplus(\Delta * x_{k_h, k_w} + bias)$
            \State $y_{k_h, k_w} = 0$
            \State \textcolor{brown}{\textit{\# loop for $N$ state dimensions}}
            \For{d = 1 to $N$}
                \State {$A^d$ = Read from HBM.}
                \State {$B^d_{k_h, k_w} =$ Read from HBM.}
                \State {$C^d_{k_h, k_w} =$ Read from HBM.}
                \State $B^d_\Delta x = B^d * \Delta * x_{k_h, k_w}$
                \State $A^d_\Delta = A^d * \Delta_x$
                \State \textcolor{brown}{\textit{\# Initialize horizontal and vertical prefix}}
                \State \textcolor{mygreen} {$P^h =$ Read $P^h_{k_h, k_w - 1}$ from HBM.}
                \State \textcolor{mygreen} {$P^v =$ Read $P^h_{k_h - 1, k_w}$ from HBM.}
                \State \textbf{Initialize} $y=0$ 
                
                \State $h^{hor,d}=parallel\_horizontal\_scan$($A^d_\Delta$, $B^d_\Delta x$,  $P^h$)
                \State \textcolor{mygreen} {Write last column of $h^{hor,d}$ as $P^h_{k_h, k_w}$ to HBM.}
                
                \State $h^d = parallel\_vertical\_scan$($A^d_\Delta$, $h^{hor,d}$, $P^v$)
                \State \textcolor{mygreen} {Write last row of $h^d$ as $P^v_{k_h, k_w}$ to HBM.}
                
                \State $y_{k_h, k_w} = y_{k_h, k_w} + C^d*h^d$
                
            \EndFor \Comment{End $N$ states}
            \State $y_{k_h, k_w} = y_{k_h, k_w} + D * x_{k_h, k_w}$
            \State {Write $y_{k_h, k_w}$ to HBM.}
        \EndFor  \Comment{End tiles}
\end{algorithmic}
\label{alg:2d_ssm_algorithm}
\end{algorithm}

\noindent\textbf{Memory access complexity}.
In the algorithm, we use $T$ to denote the height and width of a tile, and $K_H$, $K_W$ to represent the number of tiles along height and width dimensions. Thus, we have $H = K_h \times T$ and $W = K_w \times T$. We also assume $N = \mathcal{O}(H + W)$. We further use \textcolor{mygreen}{green} color to highlight the extra HBM transactions of 2DMamba compared to the vanilla Mamba. Specifically, for each tile, an extra row and column for every state dimension must be read and written to concatenate tiles. This adds an extra memory access complexity of $\mathcal{O}(NT)$ per tile. The total extra memory access complexity is therefore $K_h \times K_w \times\mathcal{O}(NT) = \mathcal{O}(N(H + W))$. Since we assume $N=\mathcal{O}(H+W)$, this simplifies to $\mathcal{O}(H\times W) = \mathcal{O}(L)$, where $L$ the the sequence length. Thus, the extra memory access complexity matches that of vanilla Mamba, ensuring the total memory access complexity remains $ \mathcal{O}(L)$. 

\noindent\textbf{Correctness}. 
The 2D scan is decomposed into a horizontal scan and a vertical scan, which are performed sequentially. That is, we first conduct the horizontal scan, and after it's done, we then conduct the vertical scan. However, each scan itself is conducted by a GPU scanner in parallel, ensuring the overall efficiency. This sequential process guarantees the correctness of the decomposition. Meanwhile, the correctness of the 1D parallel scan algorithm is elaborated by the vanilla Mamba.

\noindent\textbf{Tiling and edge cases}.
We choose two grid tile sizes: $16 \times 16$ and $32 \times 32$. Feature maps smaller than $32 \times 32$ are processed directly without tiling. Larger feature maps are tiled into $32 \times 32$ blocks. This tile size is chosen manually to balance between runtime efficiency and hardware constraints: While a larger tile size brings in higher parallelism, it comes with the cost of increased register and SRAM consumption. An excessive grid size will result in register spills, which will severely penalize performance. Following vanilla Mamba~\cite{mamba}'s practice, our current choice is the trade-off between parallelism and register spills. The tiles are processed sequentially, from top left to bottom right. This process is similar to conducting a 2D convolution over the input feature map, with kernel size equal to our tile size, and stride equal to $16$ or $32$. For input sizes that are not perfect multiples of our tile size, we pad the ``spilling'' areas with naive values $\bar{A} = 1$ and $x = 0$, which is also the strategy of the vanilla Mamba scanner. 

\noindent\textbf{Thread granularity and load balancing}.
Each feature map is processed with $64$ threads, which properly respects the $32$-thread granularity. For tile size $16 \times 16$, each thread processes a $2 \times 2$ subregion; for $32 \times 32$, each thread processes a $4 \times 4$ subregion. Accordingly, all threads process the same amount of data so \textit{load imbalances} are not an issue. 

\noindent\textbf{Difference with other hardware-optimized methods}. 
To the best of our knowledge, all currently available hardware-optimized Mamba-based methods rely on the vanilla Mamba scanner and its CUDA pipeline, which conduct 1D scans (2D input must be flattened into 1D sequence to be processed). In contrast, 2DMamba conducts 2D scans (without the need to flatten the input) and make novel modifications to the CUDA part for the best efficiency.

\begin{table*}[t]
\centering
\small
\setlength{\tabcolsep}{1.1mm}{
\begin{tabular}{llccc@{\extracolsep{3pt}}ccc@{\extracolsep{3pt}}ccc}
\toprule
&Feature size                            &
\multicolumn{3}{c}{$14\times14$} & \multicolumn{3}{c}{$56\times56$} & \multicolumn{3}{c}{$200\times200$} \\
\cline{3-5} \cline{6-8} \cline{9-11}
Scope & Method                                   & FLOPs    & Thro.    & GPU Mem.   & FLOPs    & Thro.    & GPU Mem.   & FLOPs     & Thro.    & GPU Mem.    \\ \midrule
                & Mamba (CUDA)                    & 63M     & 115        & 30 MB     & 1.0G    & 100       & 78 MB     & 12.9G     & 56        & 804 MB   \\
MIL framework   & \methodname~(Python)            & 72M      & 53        & 70 MB     & 1.2G    & 40       & 670 MB     & 14.7G     &    5     & 9110 MB  \\
                & \textbf{\methodname~(CUDA)}     & 72M     & 110        & 30 MB     & 1.2G    & 88       & 102 MB     & 14.7G     & 49        & 912 MB   \\
\bottomrule
\end{tabular}
}
\caption{Comparison of floating-point operations (FLOPs), throughput (Thro., feature maps per second), and GPU memory consumption during training. MIL frameworks are measured using $128$ dimensional feature input and the state dimension is set to $16$ for all experiments.}
\label{tab:sup:th_mem_training}
\end{table*}

\section{Mathematical derivations of 2DMamba}\label{derivation}
    We formulate 2D scanning in a manner similar to Mamba \cite{mamba} for efficient parallelism. To achieve the spatial continuity in Eq. (7), we first scan row-wise in Eq. (5) and get 
    \begin{align}
        h^{hor}_{i,j} &= \sum_{j' \leq j} \bar{A}^{(j - j')} \bar{B}x_{i,j'} \ \text{(The vanilla Mamba)} 
    \end{align}
    We then scan column-wise as Eq.(6): 
    \begin{align}
    h_{i,j} &= \bar{A}h_{i-1,j} + x^{hor}_{i,j}   \\
    &= \bar{A}h_{i-1,j} + \bar{B'}x^{hor}_{i,j} \ (\bar{B'} = I) \\
    &= \sum_{i' \leq i} \bar{A}^{(i - i')} \bar{B'}h^{hor}_{i',j} \ \text{(The vanilla Mamba)} \\
    &= \sum_{i' \leq i} \bar{A}^{(i - i')} \sum_{j' \leq j} \bar{A}^{(j - j')} \bar{B}x_{i',j'} \ (\bar{B'} = I) \\
    &= \sum_{i' \leq i} \sum_{j' \leq j} \bar{A}^{(i - i' + j - j')} \bar{B}x_{i',j'} \ \text{(Eq.(7))} 
    \end{align}

\section{Details of WSI datasets}\label{sec:supp:dataset}
    \textbf{Breast invasive carcinoma subtyping on BRACS and TCGA-BRCA.} BRACS \cite{bracs} includes 547 H\&E breast carcinoma WSIs collected from 187 patients. There are 3 classes: \textit{benign tumor} (265 slides), \textit{atypical tumor} (89 slides), or \textit{malignant tumor} (193 slides). We used the official train–validation–test split with a ratio of 395:65:87 slides. TCGA-BRCA comprise 1033 H\&E WSIs with 2 subtypes: \textit{invasive ductal carcinoma} (822 slides) and \textit{invasive lobular carcinoma} (211 slides). We follow \cite{chen2022scaling} to get the train–validation–test folds with the ratio of 841:96:96 slides.\\
    
    \noindent\textbf{Prostate cancer grading based on PANDA.} The dataset \cite{panda} consists of 10,614 digitized prostate cancer biopsies. There are 6 categories: \textit{grade 0} (2890 slides), \textit{grade 1} (2666 slides),\textit{grade 2} (1343 slides), \textit{grade 3} (1242 slides), \textit{grade 4} (1249 slides), or \textit{grade 5} (1224 slides). We label-stratified PANDA into 80:10:10 train–validation–test sets (8491:1061:1062 slides). \\
    
    \noindent\textbf{Renal cell carcinoma subtyping based on DHMC.} The dataset \cite{dhmc} include 563 H\&E WSIs collected from 485 resections and 78 biopsies. The label includes 5 types: \textit{clear cell renal cell carcinoma} (344 slides), \textit{papillary renal cell carcinoma} (101 slides) and \textit{chromophobe renal cell carcinoma} (23 slides),  \textit{renal oncocytoma} (66 slides) and \textit{benign} (29 slides). We split the dataset into 393:23:147 slides for train, validation, and test sets.\\
    
    \noindent\textbf{Non-small cell lung carcinoma subtyping on TCGA-NSCLC.} The dataset include 957 H\&E breast carcinoma WSIs, including 2 subtypes: \textit{lung adenocarcinoma} (490 slides) and \textit{lung squamous cell carcinoma} (468 slides). We follow \cite{chen2022scaling} to split the dataset into train–validation–test folds with the ratio of 785:86:87 slides.\\
    
    \noindent\textbf{Survival prediction on TCGA-KIRC, TCGA-KIRP, TCGA-LUAD, TCGA-STAD, and TCGA-UCEC.} For TCGA-KIRC (\textit{kidney renal clear cell carcinoma}), the dataset includes 498 slides, with 329 censored and 169 uncensored samples, of which 300 WSIs are used for training and 100 for validation. The TCGA-KIRP dataset (\textit{kidney renal papillary cell carcinoma}) consists of 261 slides (220 censored, 41 uncensored), with 208 WSIs for training and 53 for validation. In TCGA-LUAD (\textit{lung adenocarcinoma}), there are 455 slides (159 censored, 296 uncensored), split into 364 WSIs for training and 91 for validation. The TCGA-STAD dataset (\textit{stomach adenocarcinoma}) includes 363 slides, of which 145 are censored and 218 uncensored, divided into 290 WSIs for training and 73 for validation. Finally, the TCGA-UCEC dataset (\textit{uterine corpus endometrial carcinoma}) contains 539 slides (460 censored, 79 uncensored), with 431 WSIs allocated to training and 108 for validation. The 5-fold cross-validation splits for each dataset were derived from \cite{mambamil}.

\section{Evaluation of speed and GPU memory efficiency during training}
\label{sec:supp:speed_memory_training}
In Tab.~\ref{tab:th_mem}, we analyze speed and GPU memory efficiency during inference; here, we evaluate the same metrics during training, the floating-point operations (FLOPs), throughput, and GPU memory consumption. Since the CUDA operators are identical for both training and inference, we focus on comparing the CUDA implementation of Mamba, the Python implementation of \methodname, and the CUDA implementation of \methodname within the MIL framework, across the three input feature sizes, $14\times14$, $56\times56$, and $200\times200$. As shown in Table \ref{tab:sup:th_mem_training}, for all three sizes, with 10\% to 20\% increases of FLOPs, our method achieves throughput at approximately 90\% of vanilla Mamba while significantly outperforming the Python implementation of \methodname. In terms of GPU memory consumption, our approach incurs only a slight increase compared to vanilla Mamba while reducing memory usage by 57\% to 90\% compared to the Python implementation of \methodname. These results demonstrate that our hardware-aware 2D selective scan operator remains both fast and GPU memory-efficient during training.

\section{Additional results on natural image classification}

In Tab. \ref{tab:imagenet1k}, we apply \methodname~to the SOTA Mamba-based method on natural images VMamba~\cite{vmamba} and name it as 2DVMamba. Our results showed that 2DVMamba-T outperforms all SOTA methods. In this section, we scale 2DVMamba to its small version: 2DVMamba-S. As shown in Table \ref{supp:tab:imagenet1k}, similar to the improvements seen in the tiny version, 2DVMamba-S surpasses VMamba-S by 0.2\% with a negligible increase in FLOPs (0.1G). It also outperforms all current SOTA methods, demonstrating that our approach scales effectively to larger models.

\begin{table}[h]
\centering
\small
\begin{tabular}{lccc}
\toprule
Method                  & \#Param & FLOPs & Top-1 Acc\%   \\
\midrule
DeiT-S                  & 22M     &  4.6G & 79.8          \\
Swin-T                  & 28M     &  4.5G & 81.3          \\
Vim-S                   & 26M     &   -   & 80.3          \\
EfficientVMamba-B       & 33M     &  4.0G & 81.8          \\
LocalVMamba-T           & 26M     &  5.7G & 82.7          \\
VMamba-T                & 30M     &  4.91G & 82.6          \\
\midrule
\textbf{2DVMamba-T}     & 30M     &  4.94G & \textbf{82.8} \\
\bottomrule
DeiT-B                  & 86M     &  17.5G & 81.8          \\
Swin-S                  & 50M     &  8.7G & 83.0          \\
LocalVMamba-S           & 50M     &  11.4G & 83.7          \\
VMamba-S                & 50M     &  8.7G & 83.6          \\
\midrule
\textbf{2DVMamba-S}     & 50M     &  8.8G & \textbf{83.8} \\
\bottomrule
\end{tabular}

\caption{The top-1 accuracy (\%) of our 2DVMamba on the ImageNet-1K dataset. All images are of size $224 \times 224$.}
\label{supp:tab:imagenet1k}
\end{table}

\section{Additional ablation studies} \label{sec:supp:ablation}

\begin{table*}[ht]
    \centering
    \resizebox{1\textwidth}{!}{
    \setlength{\tabcolsep}{0.9mm}{
    \begin{tabular} {l ccc c ccc c ccc c ccc c ccc}
        \toprule
        \multirow{2}{*}{\textbf{Setting}} & \multicolumn{3}{c}{\textbf{BRACS}} && \multicolumn{3}{c}{\textbf{DHMC}} && \multicolumn{3}{c}{\textbf{PANDA}} && \multicolumn{3}{c}{\textbf{TCGA-NSCLC}}  && \multicolumn{3}{c}{\textbf{TCGA-BRCA}}\\ 
        \cline{2-4}  \cline{6-8} \cline{10-12} \cline{14-16} \cline{18-20}& 
        $Acc$ & $F1$  & $AUC$ && 
        $Acc$ & $F1$  & $AUC$ &&
        $Acc$ & $F1$  & $AUC$ &&
        $Acc$ & $F1$  & $AUC$ &&
        $Acc$ & $F1$  & $AUC$ \\            
        \midrule
        Independent $\bar{A}$ & 
        {0.7379}  & {0.6786} & 0.8857 &&  
        \textbf{0.8935}  & \textbf{0.8122} & 0.9410 && 
        0.4917  & 0.4466 & 0.8131 &&  
        \textbf{0.8851}  & \textbf{0.8851} & {0.9577} &&
        0.9333  & 0.9124 & 0.9759\\
        
        \textbf{Reused $\bar{\textbf{\textit{A}}}$}& 
        \textbf{0.7517} & \textbf{0.7045} & \textbf{0.8964} &&  
        {0.8926} & {0.8027} & \textbf{0.9468} && 
        \textbf{0.5075} & \textbf{0.4562} & \textbf{0.8184}  &&  
        \textbf{0.8851} & {0.8850} & \textbf{0.9618} &&
        \textbf{0.9458} & \textbf{0.9156} & \textbf{0.9782}\\ 
        \bottomrule
    \end{tabular}
    }
    }
    \caption{Comparison of using independent and reused parameter $\bar{A}$ in the \methodnameMIL. They achieve comparable performance.}
    \label{tab:parameter_reuse}
\end{table*}
    
\paragraph{Parameter $\bar{A}$ reuse.} In 2DMamba, $\bar{A}$ is reused for both horizontal and vertical scan to maintain the same number of parameters as the vanilla Mamba \cite{mamba}. Although we lack theoretical guarantees of optimality, our ablation studies in Tab.~\ref{tab:parameter_reuse} demonstrate that 2DMambaMIL with independent $\bar{A}$ achieves comparable performance (within $\pm{1\%}$) to 2DMambaMIL with reused $\bar{A}$ across five datasets.

\paragraph{Positional embeddings (PE).} We investigate the impact of PE in Mamba-based MIL. We compare MambaMIL, SRMambaMIL, and  \methodnameMIL with and without PE on the PANDA and TCGA-BRCA datasets. Due to the large size of WSIs, absolute PE, as in \cite{vit}, results in an excessive number of parameters for MIL models. Instead, we adopt a linear projection to map the 2D coordinates of each patch into a PE and added to the patch embeddings to integrate positional information. As shown in Table~\ref{tab:ab_pos_emb}, incorporating PE generally improves the performance of 1D Mamba-based methods, indicating the additional spatial information helps mitigate spatial discrepancies. In contrast, adding PE to our ~\methodnameMIL~reduces its performance. This decline occurs because our 2D formulation effectively integrates spatial information, making the additional PE redundant.
\begin{table}[h]
    \centering
    \small{
    \begin{tabular} {c cc cc}
        \toprule
        \multirow{2}{*}{\textbf{Method}} & \multicolumn{2}{c}{\textbf{PANDA}} & \multicolumn{2}{c}{\textbf{TCGA-BRCA}} \\ 
        \cline{2-3}  \cline{4-5} & 
        $Acc$ & $AUC$ & 
        $Acc$ & $AUC$ \\ \hline
        MambaMIL & 
        0.4679  & 0.7781 &  
        0.9333  & {0.9657} \\
        MambaMIL-PE & 
        \textbf{0.4887}  & \textbf{0.7976} & 
        \textbf{0.9292}  & \textbf{0.9705} \\
        \midrule
        SRMambaMIL &  
        0.4711  & \textbf{0.7776}  &  
        {0.9313} & {0.9657} \\ 
        SRMambaMIL-PE & 
        \textbf{0.4774}  & 0.7705 &  
        \textbf{0.9333}  & \textbf{0.9703} \\
        \midrule
        2D-MambaMIL & 
        \textbf{0.5075} & \textbf{0.8184}&  
        \textbf{0.9458} & \textbf{0.9782} \\ 
        2D-MambaMIL-PE & 
        {0.4971} & {0.8083} &  
        {0.9290} & {0.9765} \\ 
        \bottomrule
    \end{tabular}}
    \caption{The native 2D formulation of 2D-MambaMIL obtains higher performance than integrating the positional embedding (PE) into 1D-based models. Moreover, adding PE into 2D-MambaMIL consistently decreases the performance.}
    \label{tab:ab_pos_emb}
\end{table}

\paragraph{Comparison with a naive 2D method.} A naive 2D approach involves applying 1D Mamba independently to all rows and then to all columns. We compare the performance of our formulation with this naive 2D method. As shown in Table \ref{tab:naive_vs_our_method}, the naive approach is 1\%-2\% less accurate in Accuracy and AUC on the PANDA and the TCGA-BRCA datasets. Additionally, the naive method is 50\% more computationally expensive due to the padding of $14\times14$ tiles to $14\times32$ or $32\times14$, as discussed in Section \ref{subsec:method_cuda}.

\begin{table}[h]
    \centering
    \small
    \begin{tabular} {c cc cc}
        \toprule
        \multirow{2}{*}{\textbf{2D scan}} &  \multicolumn{2}{c}{\textbf{PANDA}} & \multicolumn{2}{c}{\textbf{TCGA-BRCA}} \\ 
        \cline{2-5}  & 
        $Acc$ & $AUC$ & 
        $Acc$ & $AUC$ \\         
        \hline
        Naive & 
        0.4856  &  0.8077 &
        0.9333  &  0.9760\\
        \textbf{Ours} & 
        \textbf{0.5075} & \textbf{0.8184}  &
        \textbf{0.9458} & \textbf{0.9782}\\ 
        \bottomrule
    \end{tabular}
    \caption{The comparison of naive 2D approach and our 2DMambaMIL. The naive approach applies 1D Mamba independently to all row and then all columns. Our 2DMambaMIL surpasses the naive 2D approach.}
    \label{tab:naive_vs_our_method}
\end{table}

\paragraph{2D scan order.}
    We ablate the 2D scan order of our 2DMamba by comparing two different orders: Horizontal-Vertical and Vertical-Horizontal. The results in Table \ref{tab:ab_scanning_order} show that the two scan orders of \methodname~achieve comparable performance on the PANDA and the TCGA-BRCA datasets, with the average differences of 0.5\% in accuracy and 0.6\% in AUC. This ablation shows that the scan order does not have a large influence on the performance of \methodname.

    \begin{table}[h]
        \centering
        \small
        \begin{tabular} {c cc cc}
            \toprule
            \multirow{2}{*}{\textbf{2D scan order}} &  \multicolumn{2}{c}{\textbf{PANDA}} & \multicolumn{2}{c}{\textbf{TCGA-BRCA}} \\ 
            \cline{2-5}  & 
            $Acc$ & $AUC$ & 
            $Acc$ & $AUC$ \\         
            \hline
            {Horizontal-Vertical} & 
            \textbf{0.5075} & \textbf{0.8184}&  
            \textbf{0.9458} & \textbf{0.9782} \\ 
            Vertical-Horizontal & 
            0.5001  & 0.8141 &  
            0.9427  & 0.9707 \\
            \bottomrule
        \end{tabular}
        \caption{Ablation on the 2D scan order on the PANDA and TCGA-BRCA dataset. The scan order does not have a large influence on the performance.}
        \label{tab:ab_scanning_order}
    \end{table}

\paragraph{Number of blocks.} We ablate using one, two and three 2DMamba blocks. Table \ref{tab:ablation_model_block} shows that employing one \methodname~block achieves the overall best performance. A single layer yields the highest performance in both accuracy and AUC on the TCGA-BRCA dataset and yields the best AUC with a slightly lower accuracy on the PANDA dataset.

\begin{table}[h]
    \centering
    \small
    \begin{tabular} {c cc cc}
        \toprule
        \multirow{2}{*}{\textbf{Number of blocks $U$}} &  \multicolumn{2}{c}{\textbf{PANDA}} & \multicolumn{2}{c}{\textbf{TCGA-BRCA}} \\ 
        \cline{2-5}  & 
        $Acc$ & $AUC$ & 
        $Acc$ & $AUC$ \\         
        \hline
        1 & 
        0.5075 & \textbf{0.8184} &  
        \textbf{0.9458} & \textbf{0.9782} \\ 
        2 & 
        \textbf{0.5134}  & 0.8153 &  
        0.9427  & 0.9778 \\
        3 & 
        0.5045 & 0.8178 &  
        0.9340 & 0.9558 \\ 
        \bottomrule
    \end{tabular}
    \caption{Ablation study on the number of 2DMamba blocks $U$ on the PANDA and TCGA-BRCA dataset.}
    \label{tab:ablation_model_block}
\end{table}

\paragraph{Model dimension.} We ablate the model dimensions of 2DMamba. Table \ref{tab:ablation_model_dim} depicts that, a dimension of 128 generally provides the best performance on both the PANDA and TCGA-BRCA datasets. Specifically, for the PANDA dataset, increasing the dimension to 256 or 512 slightly improves accuracy but results in a significant drop in AUC. For the TCGA-BRCA dataset, the 128-dimension model achieves the highest performance, with improvements of at least 0.9\% in accuracy and 0.7\% in AUC.

\begin{table}[h]
    \centering
    \small
    \begin{tabular} {c cc cc}
        \toprule
        \multirow{2}{*}{\textbf{Model dimension}} &  \multicolumn{2}{c}{\textbf{PANDA}} & \multicolumn{2}{c}{\textbf{TCGA-BRCA}} \\ 
        \cline{2-5}  & 
        $Acc$ & $AUC$ & 
        $Acc$ & $AUC$ \\         
        \hline
        32 & 
        0.4916 & 0.8073 &  
        0.9250 & 0.9745 \\ 
        64 & 
        0.4987  & 0.8066 &  
        0.9375  & 0.9713 \\
        128 & 
        0.5075 & \textbf{0.8184} &  
        \textbf{0.9458} & \textbf{0.9782} \\  
        256 & 
        {0.5132}  & 0.8072 &  
        0.9292 & 0.9671 \\
        512 & 
        \textbf{0.5194} & 0.8067 &  
        0.9271 & 0.9678 \\ 
        \bottomrule
    \end{tabular}
    \caption{Ablation study on the model dimensions on the PANDA and TCGA-BRCA dataset.}
    \label{tab:ablation_model_dim}
\end{table}

\paragraph{State dimension.} We ablate the state dimension $N$ of 2DMamba. Table \ref{tab:ablation_state_dim} shows that using $N=16$ provides overall the highest performance in the PANDA and TCGA-BRCA dataset. Particularly, setting $N=16$ obtains the highest AUC and the highest accuracy on TCGA-BRCA datase. On the PANDA dataset, $N=16$ obtains the highest AUC and a slightly lower accuracy, compared with $N=32$. Thus, we set $N=16$ for all our experiments.
\begin{table}[h]
    \centering
    \small
    \begin{tabular} {c cc cc}
        \toprule
        \multirow{2}{*}{\textbf{State dimension $N$}} &  \multicolumn{2}{c}{\textbf{PANDA}} & \multicolumn{2}{c}{\textbf{TCGA-BRCA}} \\ 
        \cline{2-5}  & 
        $Acc$ & $AUC$ & 
        $Acc$ & $AUC$ \\         
        \hline
        4 & 
        0.4999 & 0.8105 &  
        0.9271 & 0.9712 \\ 
        8 & 
        0.4970  & 0.8114 &  
        0.9354  & 0.9754 \\
        16 & 
        {0.5075} & \textbf{0.8184} &  
        \textbf{0.9458} & \textbf{0.9782} \\  
        32 & 
        \textbf{0.5121}  & 0.8174 &  
        0.9271  & 0.9705 \\
        64 & 
        0.5040 & 0.8096 &  
        0.9375 & 0.9773 \\ 
        \bottomrule
    \end{tabular}
    \caption{Ablation study on the state dimension $N$ on the PANDA and TCGA-BRCA dataset.}
    \label{tab:ablation_state_dim}
\end{table}

\section{Standard derivation} \label{sec:supp:std}
The metrics reported in Tab. \ref{tab:result_classification} represent the means of five runs conducted with different random seeds. The standard derivations of these metrics are provided in Tab. \ref{tab:result_classification_std}. The standard deviations of \methodname~are generally comparable to or smaller than those of other methods, demonstrating the stability of the proposed approach.

\begin{table*}[h]
    \centering
    \resizebox{1\textwidth}{!}{
    \setlength{\tabcolsep}{0.9mm}{
    \begin{tabular} {l ccc c ccc c ccc c ccc c ccc}
        \toprule
        \multirow{2}{*}{\textbf{Method}} & \multicolumn{3}{c}{\textbf{BRACS}} && \multicolumn{3}{c}{\textbf{DHMC}} && \multicolumn{3}{c}{\textbf{PANDA}} && \multicolumn{3}{c}{\textbf{TCGA-NSCLC}}  && \multicolumn{3}{c}{\textbf{TCGA-BRCA}}\\ 
        \cline{2-4}  \cline{6-8} \cline{10-12} \cline{14-16} \cline{18-20}& 
        $Acc$ & $F1$  & $AUC$ && 
        $Acc$ & $F1$  & $AUC$ &&
        $Acc$ & $F1$  & $AUC$ &&
        $Acc$ & $F1$  & $AUC$ &&
        $Acc$ & $F1$  & $AUC$ \\            
        \midrule
        AB-MIL  & 
        0.0197  & 0.0178 & 0.0199 &&  
        0.0231  & \textbf{0.0219} & 0.0273 &&
        0.0116  & 0.0139 & 0.0157 &&
        0.0104  & 0.0103 & 0.0208 &&
        0.0123  & 0.0145 & 0.0210 \\ 
        DSMIL & 
        0.0185  & \textbf{0.0165} & 0.0210 &&  
        0.0248  & 0.0220 & 0.0255 &&
        0.0118  & 0.0136 & 0.0161 &&
        0.0111  & 0.0111 & 0.0211 &&
        0.0132  & 0.0134 & 0.0201 \\
        CLAM & 
        0.0229  & 0.0189 & 0.0221 &&  
        0.0241  & 0.0258 & 0.0295 &&
        0.0141  & 0.0166 & 0.0184 &&
        0.0130  & 0.0129 & 0.0249 &&
        0.0131  & 0.0160 & 0.0268 \\
        DTFD-MIL  & 
        0.0254  & 0.0225 & 0.0255 &&  
        0.0294  & 0.0274 & 0.0345 &&
        0.0148  & 0.0179 & 0.0197 &&
        0.0133  & 0.0132 & 0.0269 &&
        0.0158  & 0.0187 & 0.0267 \\
        TransMIL  & 
        0.0262  & 0.0236 & 0.0263 &&  
        0.0310  & 0.0287 & 0.0367 &&
        0.0155  & 0.0181 & 0.0208 &&
        0.0138  & 0.0139 & 0.0274 &&
        0.0162  & 0.0190 & 0.0281 \\ 
        \midrule
        S4-MIL & 
        0.0214  & 0.0209 & 0.0194 &&  
        0.0223  & 0.0240 & 0.0246 &&
        0.0131  & 0.0139 & 0.0172 &&
        0.0116  & 0.0116 & \textbf{0.0189} &&
        0.0145  & 0.0134 & 0.0216 \\
        MambaMIL & 
        0.0179  & 0.0193 & 0.0217 &&  
        0.0229  & 0.0261 & 0.0328 &&
        0.0133  & 0.0144 & \textbf{0.0146} &&
        0.0104  & \textbf{0.0103} & 0.0249 &&
        0.0140  & \textbf{0.0132} & 0.0199 \\
        SRMambaMIL &  
        0.0189  & 0.0168 & 0.0203 &&  
        0.0232  & 0.0258 & 0.0232 &&
        \textbf{0.0107}  & 0.0147 & 0.0147 &&
        0.0117  & 0.0115 & 0.0192 &&
        0.0121  & 0.0139 & 0.0179 \\
        \midrule
        2DMambaMIL& 
        \textbf{0.0175}  & 0.0184 & \textbf{0.0173} &&  
        \textbf{0.0209}  & 0.0225 & \textbf{0.0227} &&
        \textbf{0.0107}  & \textbf{0.0132} & 0.0162 &&
        \textbf{0.0102}  & 0.0124 & 0.0205 &&
        \textbf{0.0113}  & 0.0139 & \textbf{0.0175} \\
        \bottomrule
    \end{tabular}
    }
    }
    \caption{The standard deviation of accuracy (Acc), F1 and AUC on five {WSI classification} datasets. We conducted each experiment five times using five different random seeds and reported their standard deviations. The lowest values are marked as \textbf{bold}.}
    \label{tab:result_classification_std}
\end{table*}

\section{Additional qualitative evaluation}
\label{sec:supp:qualitative}
\begin{figure*}[ht] 
          \centering
          \includegraphics[width=1\linewidth]{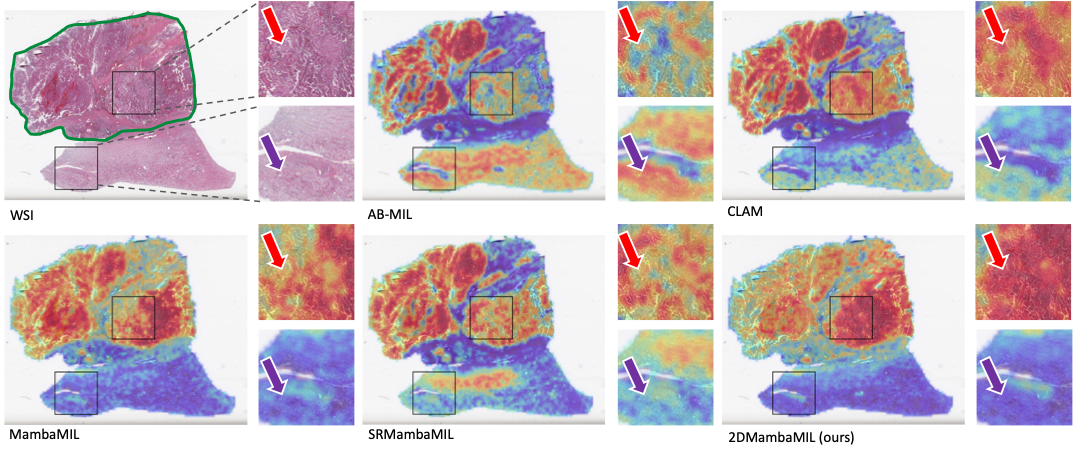}
          \caption{
          The attention visualization of \methodnameMIL~and four other methods on a TCGA-KIRP sample for \textbf{survival} analysis. Tumor regions are outlined in {\color{mygreen}green}. {\color{red}Red} arrows and {\color{violet}violet} arrows point to the highly survival relevant areas and non-tumor areas, respectively.
          2DMambaMIL and MambaMIL use tumoral and peritumoral pixels to drive the model, consistent with the heterogeneous feature of the distribution of high mortality predicting areas. By contrast, AB-MIL, CLAM, and SRMambaMIL are less specific, with high probability areas being located randomly or in insignificant structures in the non-tumoral tissue. 2DMambaMIL slightly outperforms MambaMIL in the heatmap distribution. 
          }
          \label{fig:supp_IDL_heatmap1}
    \end{figure*}

\begin{figure*}[ht] 
          \centering
          \includegraphics[width=1\linewidth]{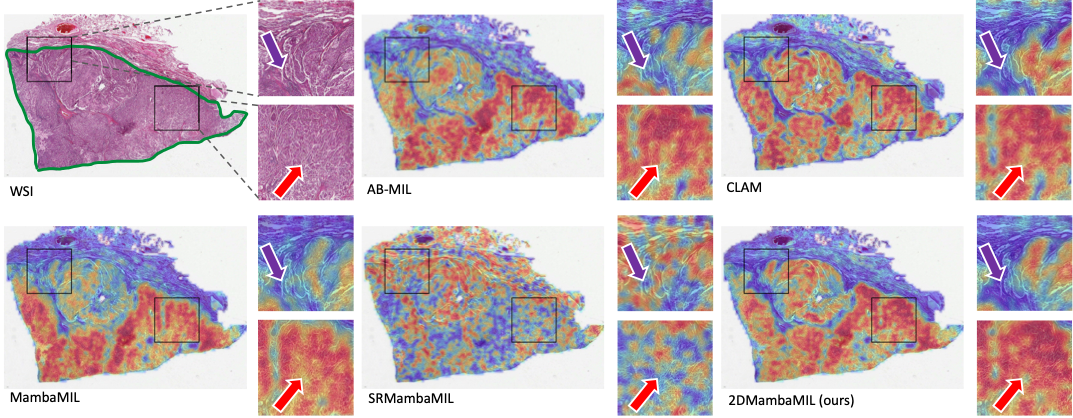}
          \caption{
          The attention visualization of \methodnameMIL~and four other methods on a TCGA-LUAD sample for \textbf{survival} analysis. Tumor regions are outlined in {\color{mygreen}green}. {\color{red}Red} arrows and {\color{violet}violet} arrows point to the highly survival relevant areas and survival irrelevant areas, respectively.
          2DMambaMIL and MambaMIL use tumoral and peritumoral pixels to drive the model, consistent with the heterogeneous feature of the distribution of high mortality predicting areas. By contrast, AB-MIL, CLAM, and SRMambaMIL are less specific, with high probability areas being located randomly or in insignificant structures in the non-tumoral tissue. 
          }
          \label{fig:supp_IDL_heatmap2}
    \end{figure*}

\begin{figure*}[ht] 
          \centering
          \includegraphics[width=1\linewidth]{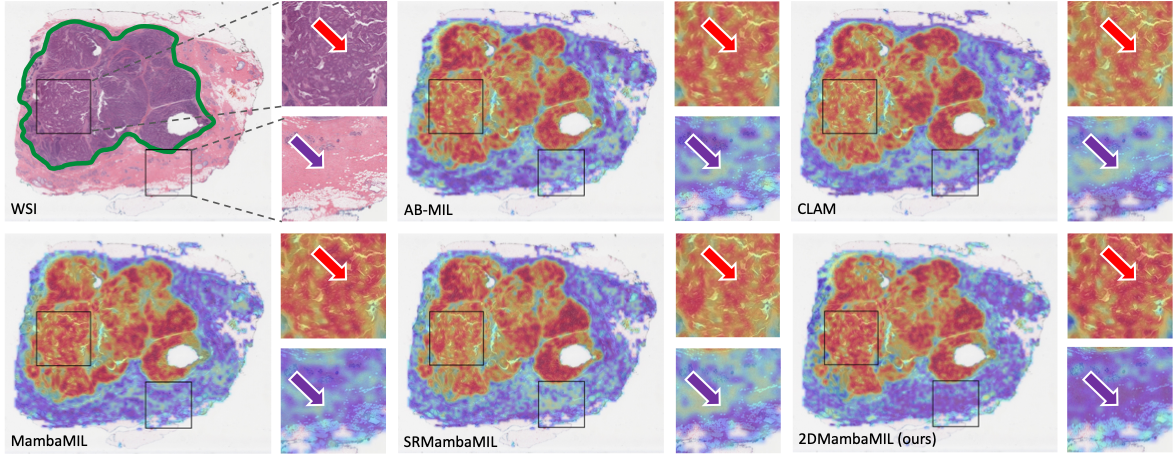}
          \caption{
          The attention visualization of \methodnameMIL~and four other methods on an IDC sample for TCGA-BRCA \textbf{sub-typing}. Tumor regions are outlined in {\color{mygreen}green}. {\color{red}Red} arrows and {\color{violet}violet} arrows point to the highly task-relevant areas and non-tumor areas, respectively.
          2DMambaMIL and MambaMIL outperform the other models in qualitative specificity, as the background non-tumor tissue contains less high-probability pixels. 2DMambaMIL slightly outperforms MambaMIL in the heatmap distribution.
          }
          \label{fig:supp_IDL_heatmap3}
    \end{figure*}

\begin{figure*}[ht] 
      \centering
      \includegraphics[width=1\linewidth]{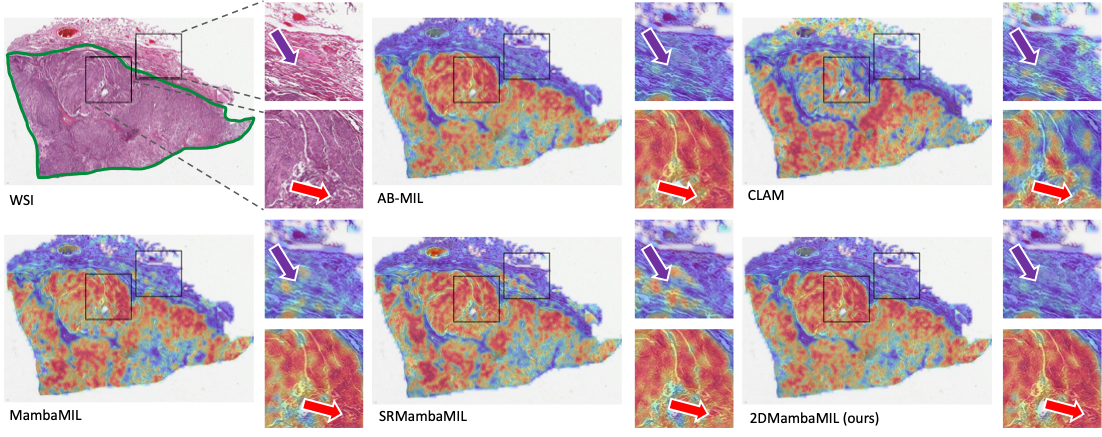}
      \caption{
      The attention visualization of \methodnameMIL~and four other methods on a LUAD sample for TCGA-NSCLC \textbf{sub-typing}. Tumor regions are outlined in {\color{mygreen}green}. {\color{red}Red} arrows and {\color{violet}violet} arrows point to the highly task-relevant areas and non-tumor areas, respectively.
      2DMambaMIL and MambaMIL outperform the other models in qualitative specificity, as the background non-tumor tissue contains fewer high-probability pixels. 2DMambaMIL slightly outperforms MambaMIL in the heatmap distribution.
      }
      \label{fig:supp_IDL_heatmap4}
    \end{figure*}

\begin{figure*}[ht] 
          \centering
          \includegraphics[width=1\linewidth]{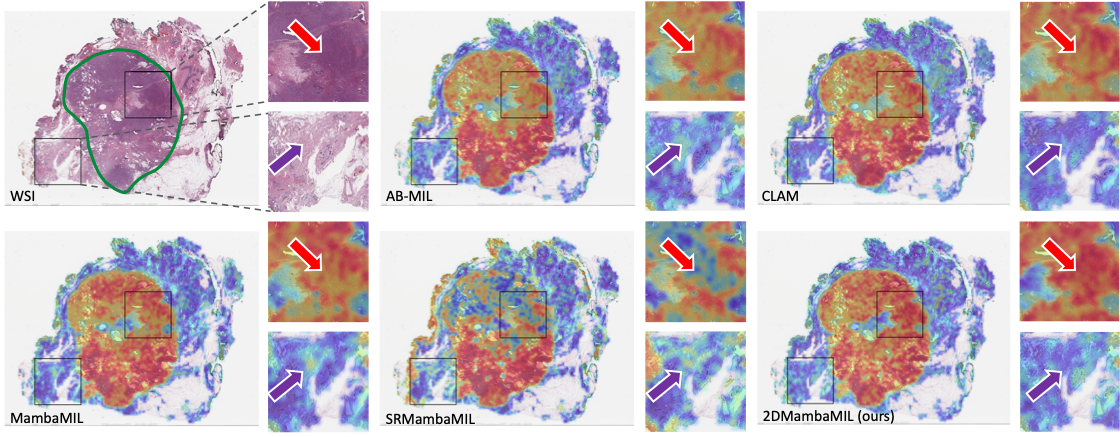}
          \caption{
          The attention visualization of \methodnameMIL~and four other methods on an ILC sample for TCGA-BRCA \textbf{sub-typing}. Tumor regions are outlined in {\color{mygreen}green}. {\color{red}Red} arrows and {\color{violet}violet} arrows point to the highly task-relevant areas and non-tumor areas, respectively.
          2DMambaMIL and MambaMIL outperform the other models in qualitative specificity, as the background non-tumor tissue contains fewer high-probability pixels. 
          }
          \label{fig:supp_IDL_heatmap5}
    \end{figure*}

\begin{figure*}[ht] 
          \centering
          \includegraphics[width=1\linewidth]{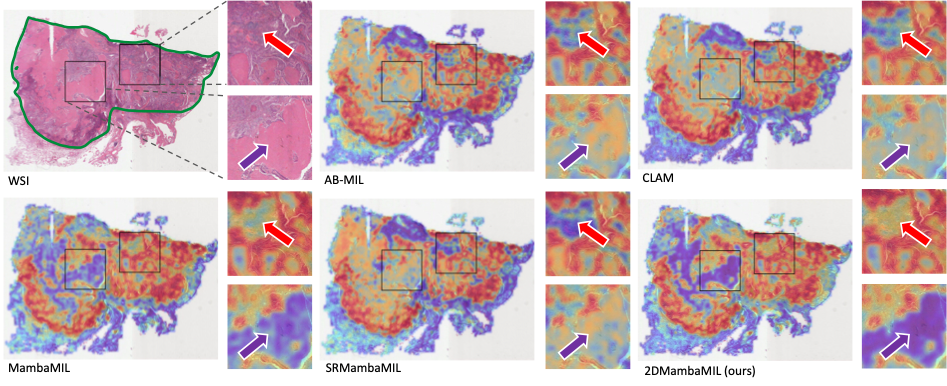}
          \caption{
          The attention visualization of \methodnameMIL~and four other methods on a LUSC sample for TCGA-NSCLC \textbf{sub-typing}. Tumor regions are outlined in {\color{mygreen}green}. {\color{red}Red} arrows and {\color{violet}violet} arrows point to the highly task-relevant areas and less task-relevant areas, respectively.
          2DMambaMIL and MambaMIL outperform the other models in qualitative specificity, as the background non-tumor tissue contains fewer high-probability pixels. 
          }
          \label{fig:supp_IDL_heatmap6}
    \end{figure*}

\begin{figure*}[ht] 
          \centering
          \includegraphics[width=1\linewidth]{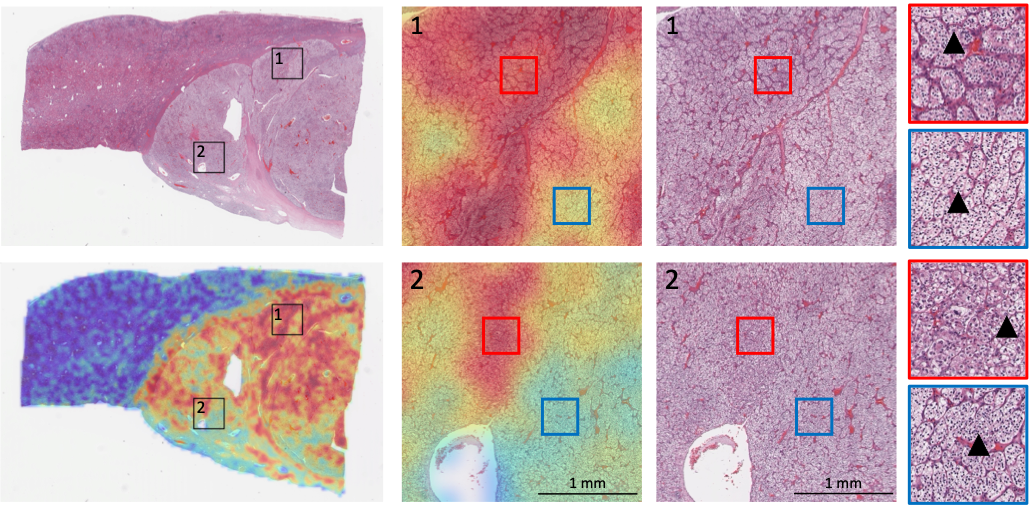}
          \caption{Two critical patches (1 and 2) of a kidney cell clear cell carcinoma sample from the TCGA-KIRC overlaid with attention heatmaps of 2DMambaMIL.
          The heatmaps of 2DMambaMIL heterogeneously show areas driving higher mortality and areas driving lower mortality. 
          Specifically, 2DMambaMIL focuses more on the \textcolor{red}{red} squares that are directly related to mortality in survival analysis and focuses less on the \textcolor{blue}{blue} squares that are less related to mortality.
          Areas in the \textcolor{red}{red} squares show features of high-grade disease (grade 2-3 pointed by black arrowheads), notably areas of tumor cells with inconspicuous nucleoli. 
          Areas in the \textcolor{blue}{blue} squares show low-grade cytological features (grade 1 pointed by black arrowheads). 
          }
          \label{fig:supp_IDL_heatmap7}
    \end{figure*}

    In addition to the heatmaps of the TCGA-KIRC sample shown in Fig. \ref{fig:heatmap}, we also analyze other samples for comprehensive qualitative evaluation.
    As shown in Fig.~\ref{fig:supp_IDL_heatmap1} to Fig.~\ref{fig:supp_IDL_heatmap6}, overall, these generated heatmaps show that 2DMambaMIL consistently generates heatmaps that are more logical than the other models. 2DMambaMIL highlights tumor features based on the task while others seem to use features from both tumor and non-tumor, showing that the model is non-specific or is tagging onto features that are not truly biologically relevant.
    For classification purposes, 2DMambaMIL heatmaps consistently highlight tumor-area pixels for classification. MambaMIL and CLAM are slightly less specific, with pixels from non-tumor areas being more often used by the model. These three models generate heatmaps that are more tumor-specific than AB-MIL and SRMambaMIL, which also highlight non-tumor features during the tumor classification task. 
    For survival prediction purposes, 2DMambaMIL also consistently used tumor areas for survival prediction while also using some pixels from the immediate tumor-adjacent areas. Interestingly, the signal detection with 2DMambaMIL was heterogeneous within the tumor, with highly- and low-attended areas of the tumor being highlighted, a feature less present with MambaMIL and CLAM, and not achieved by AB-MIL and SRMambaMIL.
    In addition, we also analyze the attention heatmap of \methodnameMIL~in high resolution patches. As shown in Fig.~\ref{fig:supp_IDL_heatmap7}, within tumor regions, our model distinguishes fine-grained regions of high and low mortality that correspond to high-grade and low-grade tumors.

\end{document}